\documentclass[journal]{IEEEtran}
\usepackage{graphics} 
\usepackage{epsfig} 
\usepackage{empheq}
\usepackage{times} 
\usepackage{amsmath} 
\usepackage{amssymb}  
\usepackage{bm}
\usepackage{color}
\usepackage{bbding} 
\usepackage{cite}
\usepackage{diagbox}
\usepackage[linesnumbered,ruled]{algorithm2e}
\usepackage{ulem} 
\usepackage{hyperref}
\usepackage{float}
\usepackage{booktabs}
\usepackage{multirow}
\usepackage{mathrsfs}
\usepackage[utf8]{inputenc}
\usepackage{subfigure}
\usepackage{graphicx}
\usepackage{pifont}
\usepackage{threeparttable}
\newcounter{RNum}
\renewcommand{\theRNum}{\arabic{RNum}}
\newcommand{\Remark}{\noindent\textbf{Remark}~\refstepcounter{RNum}\textbf{\theRNum}: }
\newcommand{\tabincell}[2]{\begin{tabular}{@{}#1@{}}#2\end{tabular}}  

\ifCLASSINFOpdf

\else

\fi

\begin{document}

\title{All-Day Object Tracking for Unmanned Aerial Vehicle}

\author{Bowen~Li,
        Changhong~Fu*,~\IEEEmembership{Member,~IEEE,}
        Fangqiang~Ding,\\
        Junjie~Ye,~\IEEEmembership{Graduate Student Member,~IEEE,}
        and Fuling Lin,~\IEEEmembership{Graduate Student Member,~IEEE}
\IEEEcompsocitemizethanks{
	\IEEEcompsocthanksitem *Corresponding author
	\protect
	\IEEEcompsocthanksitem The aurhors are with the School of Mechanical Engineering, Tongji University, 201804 Shanghai, China.
	\protect
	(Email: changhongfu@tongji.edu.cn.)
}}


\maketitle

\begin{abstract}
Visual object tracking, which is representing a major interest in image processing field, has facilitated numerous real-world applications. Among them, equipping unmanned aerial vehicle (UAV) with real-time robust visual trackers for all-day aerial maneuver, is currently attracting incremental attention and has remarkably broadened the scope of applications of object tracking. However, prior tracking methods have merely focused on robust tracking in the well-illuminated scenes, while ignoring trackers' capabilities to be deployed in the dark. In darkness, the conditions can be more complex and harsh, easily posing inferior robust tracking or even tracking failure. To this end, this work proposed a novel discriminative correlation filter-based tracker with illumination adaptive and anti-dark capability, namely ADTrack. ADTrack firstly exploits image illuminance information to enable adaptability of the model to the given light condition. Then, by virtue of an efficient and effective image enhancer, ADTrack carries out image pretreatment, where a target-aware mask is generated. Benefiting from the mask, ADTrack aims to solve a dual regression problem where dual filters, \textit{i.e.}, the context filter and target-focused filter, are trained with mutual constraint. Thus ADTrack is able to maintain continuously favorable performance in all-day conditions. Besides, this work also constructed one UAV nighttime tracking benchmark UAVDark135, comprising of more than 125k manually annotated frames, which is also very first UAV dark tracking benchmark. Exhaustive experiments are extended on authoritative daytime benchmarks, \textit{i.e.}, UAV123@10fps, DTB70, and the newly built dark benchmark UAVDark135. Our results have validated the superiority of ADTrack in both bright and dark conditions compared with other arts. Meanwhile, ADTrack realizes a real-time speed of over 30 frames/s on a single CPU, remarkably ensuring real-world UAV object tracking under all-day scenes.
\end{abstract}


\IEEEpeerreviewmaketitle

\section{Introduction}

\IEEEPARstart{S}{tanding} as one of the hotspots in image processing field, visual object tracking aims at estimating the location and scale of an initial given object in the subsequent frames of an image sequence. Applying such flourishing approach onboard unmanned aerial vehicle (UAV) has enabled extensive applications in practice, \textit{e.g.}, path optimization and planning \cite{xu2017SP}, disaster response \cite{yuan2015ICUAS}, target following \cite{Vanegas2017AC}, and autonomous landing \cite{Lin2017AR}. Specifically, transmission-line inspection \cite{Bian2018IROS}, collision avoidance \cite{Baca2018IROS}, and aircraft tracking \cite{Fu2014ICRA}, often need around-the-clock operation. 

\begin{figure}[!t]
	\centering
	\includegraphics[width=1\columnwidth]{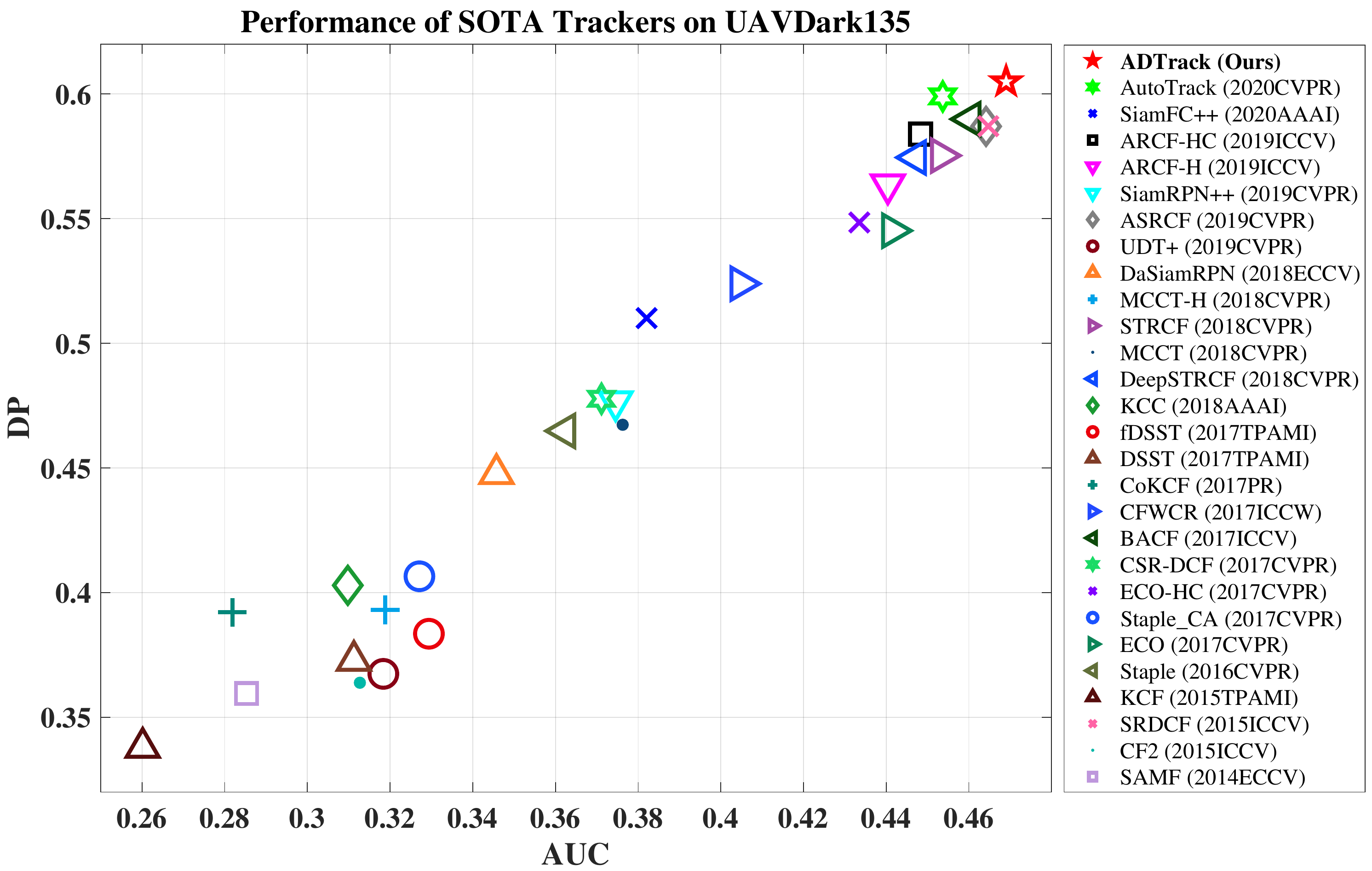}
	\vspace{-0.6cm}
	\caption{Performance comparison of SOTA trackers on the newly constructed UAVDark135, where tracking scenes are generally much darker and more onerous. Clearly, ADTrack outperforms the other trackers in both distance precision (DP), and area under curve (AUC), maintaining favorable robustness even in harsh dark conditions.}
	\label{fig:Star}
\end{figure}

\begin{figure*}[!t]
	\centering
	\includegraphics[width=2.04\columnwidth]{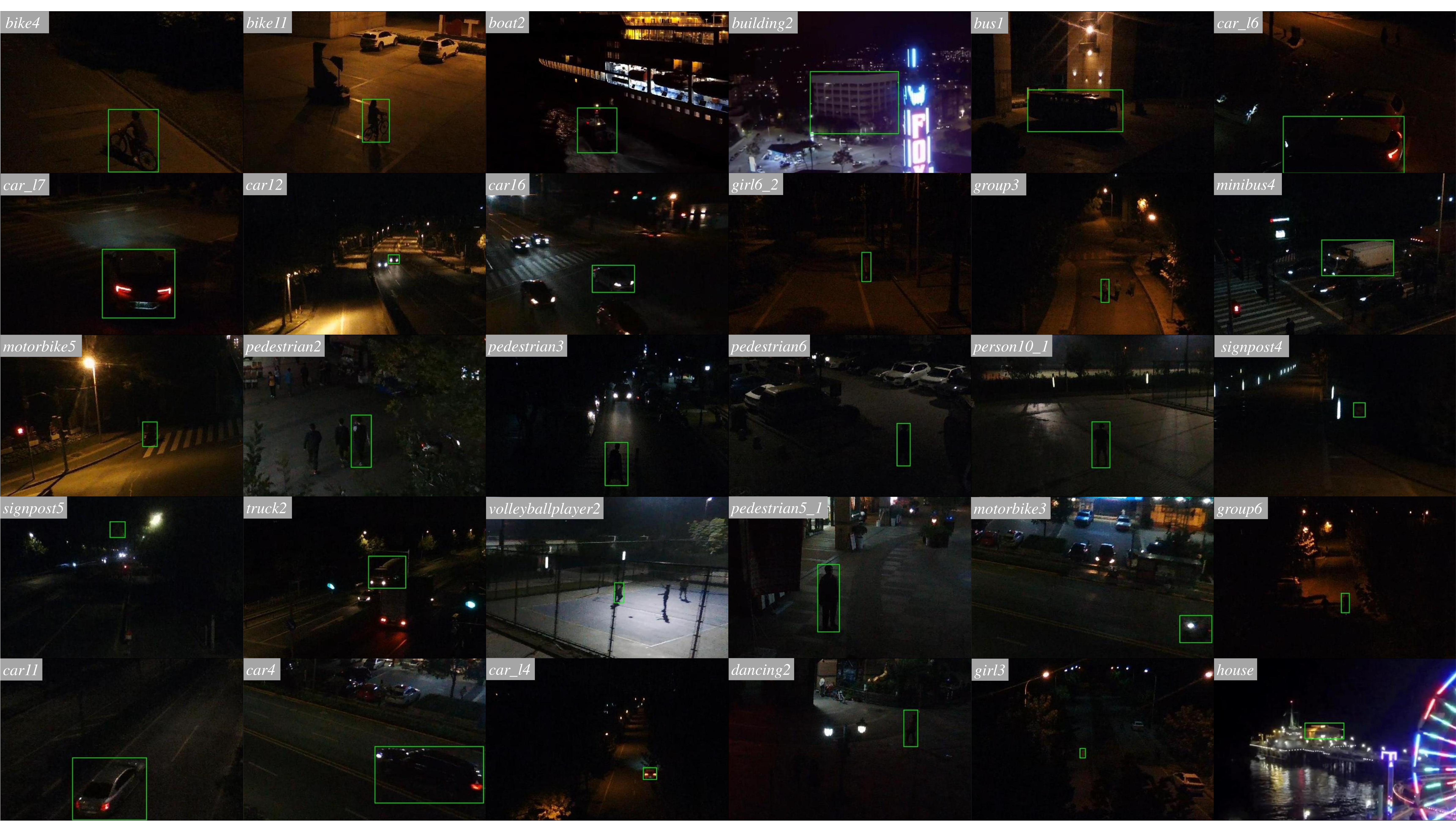}
	\vspace{-0.6cm}
	\caption{The first frames of representative scenes in newly constructed UAVDark135. Here, target ground-truths are marked out by \textbf{\textcolor{green}{green}} boxes and sequence names are located at the top left corner of the images. Dark special challenges like objects' unreliable color feature and objects' merging into the dark can be seen clearly. }
	\vspace{-0.6cm}
	\label{fig:UAVDark135}
\end{figure*}
Unfortunately, state-of-the-art (SOTA) trackers \cite{Li2020CVPR,Huang2019ICCV,Galoogahi2017ICCV,danelljan2017TPAMI,Wang2020TIP,Han2020TIP,wang2019CVPR,Xu2020AAAI,Guo2017ICCV} only focus on tracking in bright environment, where external illumination condition is favorable and the inline texture as well as outline of the object is representatvie. When the night falls, despite that the content of the scene is discernible, the visual quality of images captured is barely satisfactory, hurting the performance of methods
that are primarily designed for object tracking with high-visibility inputs.
Compared with common tracking scenes, tracking in the dark onboard UAV confronts special undesirable hardships such as:
\begin{itemize}
	\item Color distortion: Since the objects in the dark are not well-illuminated, very little light is reflected on their surfaces, making them nearly all-dark. In this case, the objects' color is distorted. Without representative color features, the discriminative ability of the tracker can decrease in a notable margin.
	
	\item Low visibility: When the object enters dimmer region in the dark, like the shadow, it can merge into the dark background, ending up in low visibility. Such scenes set barrier to trackers' robust localization and precise scale estimation of the object.
	
	\item High-level noise: Images captured by UAV at night inevitably contains random noise, which may distract trackers, resulting in inferior robust tracking.
	
	\item Limited computation: For cost issue and scarce battery power, UAV generally carries a single CPU as its computation platform. While in order to handle tough darkness, more modules are needed to maintain tracking robustness, making real-time processing even more arduous.
\end{itemize}

Though the outstanding trackers \cite{Li2020ICRA,Li2020CVPR,Huang2019ICCV,Galoogahi2017ICCV,danelljan2017TPAMI,Bo2019CVPR,li2019CVPR,wang2019CVPR,Xu2020AAAI,Guo2017ICCV} have achieved promising performance in well-illuminated scenes. The deep trackers \cite{Bo2019CVPR,li2019CVPR,wang2019CVPR,Xu2020AAAI,Guo2017ICCV} on the one hand introduce too much computation burden to be deployed on a single CPU for real-time tracking. On the other, the discriminative ability of deep semantic features were proved to drop significantly in our experiment since the off-the-shelf network is trained by bright images. While the brilliant discriminative correlation filter (CF)-based trackers \cite{Li2020ICRA,Li2020CVPR,Huang2019ICCV,Galoogahi2017ICCV,danelljan2017TPAMI}, which utilize online learned handcrafted features, easily lose accuracy and robustness under low-light nighttime scenes due to the challenges mentioned above. In Fig.~\ref{fig:Star}, both deep and handcrafted CF-based methods fails to meet our expectation. Prior work contributed very few to robust tracking in the dark, while there is an urgent need to develop to broaden the service life and usage scenarios of UAV.

In this regard, this work proposes a novel and pioneer tracker with \textit{illuminance adaptive} and \textit{anti-dark} function (ADTrack) to achieve robust all-day real-time UAV tracking. Fig.~\ref{fig:Star} exhibits the superiority of proposed ADTrack against other arts in nighttime UAV tracking scene.

To be specific, ADTrack explores image illumination processing methods \cite{Reinhard2002ACM,Ahn2013ICCE} and proposes more innovative module to be embedded into efficient robust CF-based tracker \cite{Galoogahi2017ICCV}. To achieve robust tracking under 2 distinct light condition, \textit{i.e.}, day and night,  ADTrack firstly exploits image illumination \cite{Reinhard2002ACM} to detect and adapt to the condition, which is inspired by human visual system and proves to be effective and efficient. Then, a fast enhancer \cite{Ahn2013ICCE} generates appropriate image samples according to the detection result for training.

Surprisingly, we found the image illumination map in \cite{Ahn2013ICCE} can be utilized to obtain a target-aware mask. By virtue of the mask, ADTrack solves a dual regression model to train target-focused and context filters with mutual constraint. During detection phase, dual responses, \textit{i.e.}, target-focused and context response maps, are fused using weight sum to achieve more precise object localization. With the proposed dual filters, ADTrack proves to be more robust in around-the-clock UAV tracking scenes.

Besides, to the best of our knowledge, there exists no dark UAV tracking benchmark for large-scale evaluation in literature. Thus, this work also builds the very first UAV dark tracking benchmark, \textit{i.e.}, UAVDark135. UAVDark135 consists of totally 135 sequences, most of which were newly shot by a standard UAV at night, including more than 125k manually annotated frames. The benchmark covers a wide range of scenes, \textit{e.g.}, road, ocean, street, highway, and lakeside, including a large number of objects, such as person, car, building, athlete, truck, and bike. Fig.~\ref{fig:UAVDark135} displays representative scenes in UAVDark135, where dark special challenges are distinct.

Contributions\footnote{The source code of the proposed ADTrack and newly constructed benchmark UAVDark135 are located at \url{https://github.com/vision4robotics/ADTrack_v2}.} of this work can be summarized as:
\begin{itemize}
	\item This work proposes a novel tracker with illumination adaptive and anti-dark function (ADTrack).
	
	\item The proposed ADTrack exploits image illumination to acquire target-aware mask, which can be creatively utilized to solve a dual regression.
	
	\item This work constructed the very pioneer UAV dark tracking benchmark UAVDark135 to perform large-scale evaluation. 
	
	\item Exhaustive experiments have been conducted on two authoritative daytime benchmark UAV123@10fps, DTB70 and the newly built nighttime benchmark UAVDark135 to validate the surprising ability of proposed ADTrack in around-the-clock tracking performance.
\end{itemize}

The remainder of this work is organized as follows. Section~\ref{Sec:2} summarizes related work about image enhancing approaches, CF-based tracking methods, and target-aware tracking approaches. Section~\ref{Sec:3} elaborately interprets the proposed ADTrack, which can be epitomized as 4 modules, respectively, illumination adaptation, pretreatment, filter training, and object detection. Section~\ref{Sec:4} gives a thorough introduction of the newly built UAVDark135, including its statistics, platform, annotation, and attributes. Section~\ref{Sec:5} exhibits comprehensive experimental evaluation on various benchmarks, \textit{i.e.}, UAV123@10fps \cite{Mueller2016ECCV}, DTB70 \cite{Li2017AAAI}, and UAVDark135, where the superiority of ADTrack is apparent. Finally, Section~\ref{Sec:6} gives an integrated conclusion of the full article.

\section{Related Work}\label{Sec:2}
\subsection{Low-Light Image Enhancers}
Existing SOTA enhancers can be generally divided into 2 categories, \textit{i.e.}, learning-based and model-based. learning-based methods aim at training an end-to-end network specialized for domain transformation with paired images \cite{Wang2018gladnet,Chen2018CVPR,Ren2019TIP}. To name a few, C. Chen \textit{et al.} \cite{Chen2018CVPR} carefully designed an encoder-decoder structured network and trained it with paired short-exposure low-light images and corresponding long-exposure images, which can generate high-quality enhanced images. W. Ren \textit{et al.} \cite{Ren2019TIP} proposed a deep hybrid network made up of distinct content stream and edge stream, which can handle the degraded images captured at night. Such methods are usually carried out on GPU due to their huge amount of calculation brought by convolution layers, which cannot realize real-time processing on a single CPU for tracking onboard UAV.

Model-based methods \cite{Li2018TIP,Meylan2006TIP,Ahn2013ICCE,Rahman1996ICIP} are in view of the famous retinex theory \cite{Land1977SA}, which need no off-line training. For instance, M. Li \textit{et al.} \cite{Li2018TIP} creatively proposed to estimate a noise map based on tradition retinex model, which is able to obtain de-noised enhanced images. Z. Rahman \textit{et al.} \cite{Rahman1996ICIP} replaced the original logarithm function in multi-scale retinex model with a sigmoid function which can suppress noise speckles in extreme low light areas.

Specially, the enhancer \cite{Ahn2013ICCE} proves to be both efficient and effective. This work introduces it into the UAV tracking structure and constructs an illumination adaptive anti-dark tracker. Different from roughly applying the enhancer to preprocess the frames, ADTrack achieves light condition awareness and dual regression with the in-depth integration of image enhancement and visual tracking.

\subsection{CF-Based Tracking Methods}
The key idea of CF-based tracking methods is to train a discriminative correlation filter using current image samples and utilize the trained filter to search for the object in the coming frame \cite{Henriques2015TPAMI,Danelljan2015ICCV,Galoogahi2017ICCV,Li2018CVPR,danelljan2017TPAMI,Huang2019ICCV,Li2020CVPR,bertinetto2016CVPR}. J. F. Henriques \textit{et al.} introduced circular sample matrix, ridge regression and kernel correlation in the KCF tracker \cite{Henriques2015TPAMI}, considered as the foundation of all CF-based trackers. M. Danelljan \textit{et al.} proposed scale filter in the DSST tracker \cite{danelljan2017TPAMI}, settling scale estimation efficiently. H. K. Galoogahi \textit{et al.} put forward cropping matrix in the training regression equation \cite{Galoogahi2017ICCV}, not only immensely expanding the real negative samples but also alleviating boundary effect.

Generally speaking, the key enablers of the superiority of CF type methods involve: $1$) its simplicity of calculation by discrete Fourier transformation as well as a myriad of implicitly circular shift samples generated in this duration, $2$) its online learning schemes which make the filter maintain satisfying robustness in various scenarios. UAV tracking, owing to its limited computation resource and wide application, is right the field where CF-based trackers can shine brilliantly \cite{Fu2020TGRS,Li2020ICRA,Ding2020IROS,Huang2019ICCV,Li2020CVPR}. To be specific, Z. Huang \textit{et al.} exploited response map aberrance and proposed the ARCF tracker \cite{Huang2019ICCV}, which ensures high robustness under volatile tracking distractors. The AutoTrack tracker \cite{Li2020CVPR} proposed by Y. Li \textit{et al.} aimed at the onerous mode adjustment procedure and adapted to various given condition automatically.

Though the trackers mentioned above can strike a balance between accuracy and speed in common bright environment, they lose robustness and accuracy in the dark, when the light condition becomes abominable. To realize all-day real-time UAV tracking, this work proposed ADTrack, which can adapt to the given light condition and maintain predominant robustness even in the terrible darkness.

\subsection{Target-Aware CF-Based Tracking}
Target-aware mask, which is a foreground matrix, concentrating the filter's attention on pixels more possible included within the object outline. Such strategy can raise the importance of features indeed represent the object characteristics. M. Danelljan \textit{et al.} proposed a spatial punishment term in the SRDCF tracker \cite{Danelljan2015ICCV}, which can be considered as a foreground mask, making the filter learn center region more to alleviate boundary effect. A. Lukezic \textit{et al.} improved the fixed spatial regularization term in the SRDCF tracker \cite{Danelljan2015ICCV} into alterable reliability mask in the CSR-DCF tracker \cite{Lukezic2017CVPR}, which is based on Bayes principle. Recently, saliency-aware methods are also introduced \cite{Fu2020TGRS,Feng2019TIP}. Specifically, C. Fu \textit{et al.} \cite{Fu2020TGRS} creatively utilized an efficient saliency detection method to generate effective mask, which raised the robustness tracker onboard UAV tremendously.

\begin{figure*}[!t]
	\centering
	\includegraphics[width=2.04\columnwidth]{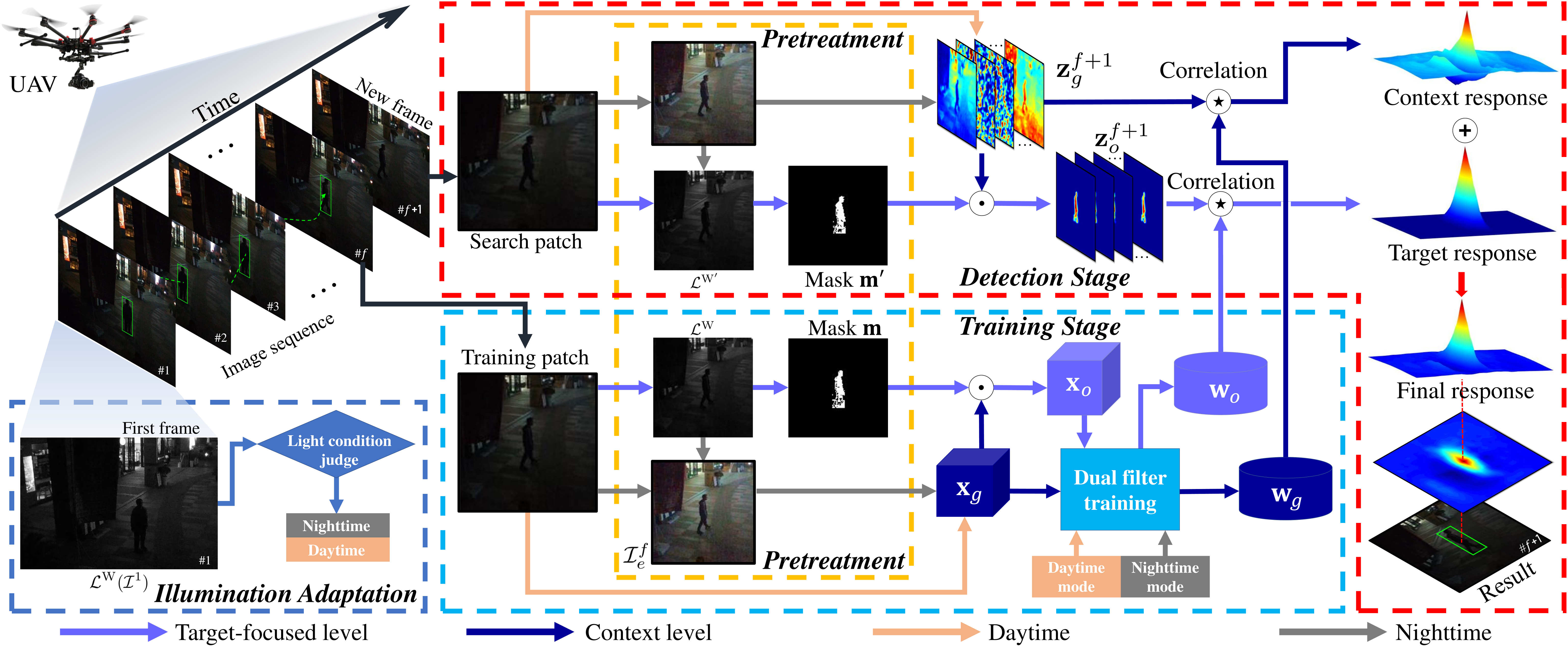}
	\vspace{-0.3cm}
	\caption{Pipeline of ADTrack. The proposed tracker contains four stages, \textit{i.e.}, illumination adaptation, pretreatment, training, and detection stages, which are marked out by boxes in different colors. In well-illuminated daytime and dim nighttime, ADTrack is able to adjust its tracking modules automatically, according to light condition judgment in illumination adaptation stage. In training and detection stage, ADTrack adopts dual filter training and dual response fusion, respectively, target-focused level and context level, as is shown in different routes.}
	\label{fig:main}
	\vspace{-0.3cm}
\end{figure*}

Despite that the methods mentioned can improve tracking performance, they have two obvious drawbacks. Firstly, it is hard to obtain valid alterable masks like \cite{Lukezic2017CVPR}, \cite{Feng2019TIP}, and \cite{Fu2020TGRS} in the dark, since nighttime images lack sufficient information. Secondly, the aforementioned trackers directly embed the masks into the CF in training process as a regularization term, assigning higher weights to the predicted target region in the filter. When an invalid mask generates, wrong pixels that odd with actual target region of the CF will possess higher importance, easily leading to tracking failure. 

Totally different from the above, ADTrack exploits image illumination information to obtain effective masks in arbitrary light condition. By virtue of the target-aware mask, ADTrack proposes a dual regression, where target-focused and context filters are trained with mutual constriction. In this way, both background and target information are learned and utilized, greatly improving tracking robustness.

\section{Illumination Adaptive and Anti-Dark Object Tracking}\label{Sec:3}

The pipeline of the proposed ADTrack can be divided into four stages, \textit{i.e.}, illumination adaptive stage, pretreatment stage, training stage, and detection stage. As is clearly exhibited in Fig.~\ref{fig:main}, for a given dark tracking scene, ADTrack firstly implements an illumination adaptation decider in the first frame captured by UAV to judge whether it is at daytime or nighttime. The determined outcome can enable mode switching automatically. In the subsequent frame $f$, ADTrack carries out pretreament stage with the illumination judgment result, where appropriate samples (from original image or enhanced image) and target-aware mask are obtained simultaneously. Then in training stage, dual filters are jointly trained by focusing on both context and target appearance. As next frame $f+1$ comes, the trained filters generate dual response maps which are fused to obtain the final response for target localization as detection stage.

\subsection{Illumination Adaptation}\label{IA}
To realize all-day adaptation, we consider an effective illumination expression algorithm, which transforms complex image illuminance information into a simple constant.
For a given RGB image $\mathcal{I}\in\mathbb{R}^{w\times h\times 3}$, the pixel-level world illumination value $\mathcal{L}^{\mathrm{W}}(\mathcal{I})$ is firstly computed as:
\begin{equation}\label{eqn:1}
\mathcal{L}^{\mathrm{W}}(x,y,\mathcal{I})=\sum_{\mathrm{m}}\alpha_{\mathrm{m}}\Psi_{\mathrm{m}}(\mathcal{I}(x,y)),~\mathrm{m}\in\{\mathrm{R,G,B}\}~,
\end{equation}
where $\Psi_{\mathrm{m}}(\mathcal{I}(x,y))$ denotes the intensity value of image $\mathcal{I}$ at coordinate $(x,y)$ in color channel $\mathrm{m}$, \textit{e.g.,} $\Psi_{\mathrm{G}}(\mathcal{I}(x,y))$ denotes the value in green channel. The channel weight parameters $\alpha_\mathrm{R},\alpha_\mathrm{G},\alpha_\mathrm{B}$ meet $\alpha_\mathrm{R}+\alpha_\mathrm{G}+\alpha_\mathrm{B}=1$. Then, the log-average luminance $\tilde{\mathcal{L}}^{W}(\mathcal{I})$ is given as in \cite{Reinhard2002ACM}:
\begin{equation}\label{eqn:2}
\tilde{\mathcal{L}}^{\mathrm{W}}(\mathcal{I})={\mathrm{exp}}\Big(\frac{1}{wh}\sum_{x,y}\mathrm{log}(\delta+\mathcal{L}^{\mathrm{W}}(x,y,\mathcal{I}))\Big)~,
\end{equation}
where $\delta$ is a small value, to avoid zero value.

\Remark Our experiment has proved that the log-average luminance $\tilde{\mathcal{L}}^{W}(\mathcal{I})$ can express the light condition of image $\mathcal{I}$ effectively.

Different light condition, \textit{e.g.}, in the dark or in daytime, the log-average luminance varies largely. Thus, ADTrack introduced a threshold $\tau$ for illumination judgment as:
\begin{equation}\label{eqn:3}
S(\mathcal{I})=\left\{
\begin{array}{rcl}
1 & & {\tilde{\mathcal{L}}^{W}(\mathcal{I}) < \tau}\\
0 & & {\tilde{\mathcal{L}}^{W}(\mathcal{I}) \ge \tau}
\end{array}\right.~,
\end{equation}
where $S(\mathcal{L})$ can be seen as the night identifier, $S(\mathcal{I})=1$ indicates that image $\mathcal{I}$ is a night image.

\Remark To test the validity of the above night decider and determine a proper $\tau$, this work selected first frames in benchmark UAV123@10fps \cite{Mueller2016ECCV} as daytime test samples and newly constructed benchmark UAVDark135 as nighttime test samples. The deciding success rate of different thresholds $\tau$ are shown in TABLE~\ref{tab:ildecider}, where the decider can achieve a surprising result of over 99\%.

\Remark During UAV tracking, ADTrack implements illumination adaptation decider in the first frame of a given sequence, then adjusts its mode and following pretreatment stage automatically according to the judgment outcome $S(\mathcal{I})$ in Eq.~(\ref{eqn:3}).
\begin{table}[t]
	\centering
	\setlength{\tabcolsep}{1.2mm}
	\fontsize{8}{8}\selectfont
	\begin{threeparttable}
		\caption{Success rates of proposed illuminance decider with different thresholds $\tau$. Clearly, the results can surprisingly achieve over 99\%, enabling effective night judgment.}
		\vspace{0.08cm}
		\begin{tabular}{ccccccccc}
			\toprule[1.5pt]
			Thresholds $\tau$ &0.12
			& 0.13 & 0.14 & 0.15 & 0.16&0.17&0.18&0.19 \\
			\midrule
			Success rate &0.979
			& 0.983 & 0.987 & \textbf{0.991} & 0.987&0.987 &\textbf{0.991} &0.983\\
			\bottomrule[1.5pt]			
		\end{tabular}\label{tab:ildecider}
		\vspace{-0.1cm}
	\end{threeparttable}
\end{table}

\subsection{Pretreatment}\label{pre}

The pretreatment stage can achieve two purposes. Firstly, for determined night sequences, ADTrack adopts an efficient and effective image enhancer \cite{Ahn2013ICCE} to obtain enhanced images for the subsequent training and detection stages. Secondly, the target-aware mask is acquired by virtue of image illuminance information, so that dual filters learning in the training process and dual response maps generation in the detection process can be realized.

\Remark To our excitement, the two purposes can be both based on world illumination value $\mathcal{L}^{\mathrm{W}}(x,y,\mathcal{I})$ in Eq.~(\ref{eqn:1}) and log-average luminance $\tilde{\mathcal{L}}^{W}(\mathcal{I})$ in Eq.~(\ref{eqn:2}).

To realize image enhancing, the global adaptation factor $\mathcal{L}_{\mathrm{g}}(x,y,\mathcal{I})$, which is based on the original world illumination map, is firstly calculated as in \cite{Ahn2013ICCE}:
\begin{equation}\label{eqn:4}
\mathcal{L}_{\mathrm{g}}(x,y,\mathcal{I})=\frac{\mathrm{log}(\mathcal{L}^{\mathrm{W}}(x,y,\mathcal{I})/\tilde{\mathcal{L}}^{\mathrm{W}}(\mathcal{I})+1)}{\mathrm{log}(\mathcal{L}^{\mathrm{W}}_{\mathrm{max}}(\mathcal{I})/\tilde{\mathcal{L}}^{\mathrm{W}}(\mathcal{I})+1)}~,
\end{equation}
where $\mathcal{L}^{\mathrm{W}}_{ \mathrm{max}}(\mathcal{I})=\mathrm{\mathrm{max}}(\mathcal{L}^{\mathrm{W}}(x,y,\mathcal{I}))$. The calculated factor can be referred to change the pixel value in three intensity channels of each pixel to realize image enhancement as:
\begin{equation}\label{eqn:5}
\Psi_{\mathrm{m}}(\mathcal{I}_\mathrm{e}(x,y))=\Psi_{\mathrm{m}}(\mathcal{I}(x,y))\cdot\frac{\mathcal{L}_{\mathrm{g}}(x,y,\mathcal{I})}{\mathcal{L}^{\mathrm{W}}(x,y,\mathcal{I})}~,
\end{equation}
where $\mathcal{I}_{\mathrm{e}}$ denotes the enhanced image based on original $\mathcal{I}$. Since $\mathcal{L}_{\mathrm{g}}(x,y,\mathcal{I})$ varies in different regions that process different illumination, Eq.~(\ref{eqn:5}) can readjust the brightness of the whole image while keeping the proportion of the three color channels a constant, \textit{i.e.}, the image color unchanged.
\Remark The aforementioned strategy is merely the fast first stage of \cite{Ahn2013ICCE}, Eq.~(\ref{eqn:5}) shows its efficacy for image enhancing.

For target-aware mask generation, ADTrack considers the illuminance change $\bm{\Theta}_{\mathcal{L}}(\mathcal{I})$ after enhancement, which can be written as:
\begin{equation}\label{eqn:6}
\begin{split}
\bm{\Theta}_{\mathcal{L}}(\mathcal{I})&=\mathcal{L}^{\mathrm{W}}(\mathcal{I})-\mathcal{L}^{\mathrm{W}}(\mathcal{I}_\mathrm{e})\\
&=\frac{\mathcal{L}^{\mathrm{W}}(x,y,\mathcal{I})-\mathrm{log}\Big(\frac{\mathcal{L}^{\mathrm{W}}(x,y,\mathcal{I})}{\tilde{\mathcal{L}^{\mathrm{W}}}(\mathcal{I})+1}\Big)}{\mathrm{log}(\mathcal{L}^{\mathrm{W}}_{ \mathrm{max}}(\mathcal{I})/\tilde{\mathcal{L}}^{\mathrm{w}}(\mathcal{I})+1)}~.
\end{split}
\end{equation}
\\
\Remark The illumination change $\bm{\Theta}_{\mathcal{L}}(\mathcal{I})(x,y)$ of pixel $(x,y)$ in a given image depends only on the first 2 terms in the numerator of Eq.~(\ref{eqn:6}), \textit{i.e.}, $\mathcal{L}^{\mathrm{W}}(x,y,\mathcal{I})-\rm{log}(\mathcal{L}^{\mathrm{W}}(x,y,\mathcal{I}))$. Since $\mathcal{L}^{\mathrm{W}}(x,y,\mathcal{I})\in[0,1]$, the value of $\bm{\Theta}_{\mathcal{L}}(\mathcal{I})$ apparently varies according to the original illumination $\mathcal{L}^{\mathrm{W}}(x,y,\mathcal{I})$. Assume that different objects' illumination are different under similar light condition in an image due to their various reflectivity. By virtue of Eq.~(\ref{eqn:6}), the class of pixels can be indicated as the target or the context. 

To be specific, consider that the target is located at the center part, the average value $\mu$ and standard deviation $\sigma$ of the center region of $\mathbf{\Theta}_{\mathcal{L}}$ are computed. Following a three-sigma criterion in statistics, which reflects the probability distribution characteristics of samples, pixels in the range $\mu\pm3\sigma$ are considered targets while others are the context. Then, a binary mask $\mathbf{m}_r$ is generated as:
\begin{equation}
\mathbf{m}_r(x,y)=\left\{
\begin{array}{rcl}
1 & & {\mu-3\sigma \le \mathbf{\Theta}_{\mathcal{L}}(x,y) \le \mu+3\sigma}\\
0 & & {\mathrm{else}}
\end{array}
\right.~.
\end{equation}

\begin{figure}[!b]
	\centering
	\setlength{\abovecaptionskip}{-0.3cm}
	\includegraphics[width=1\columnwidth]{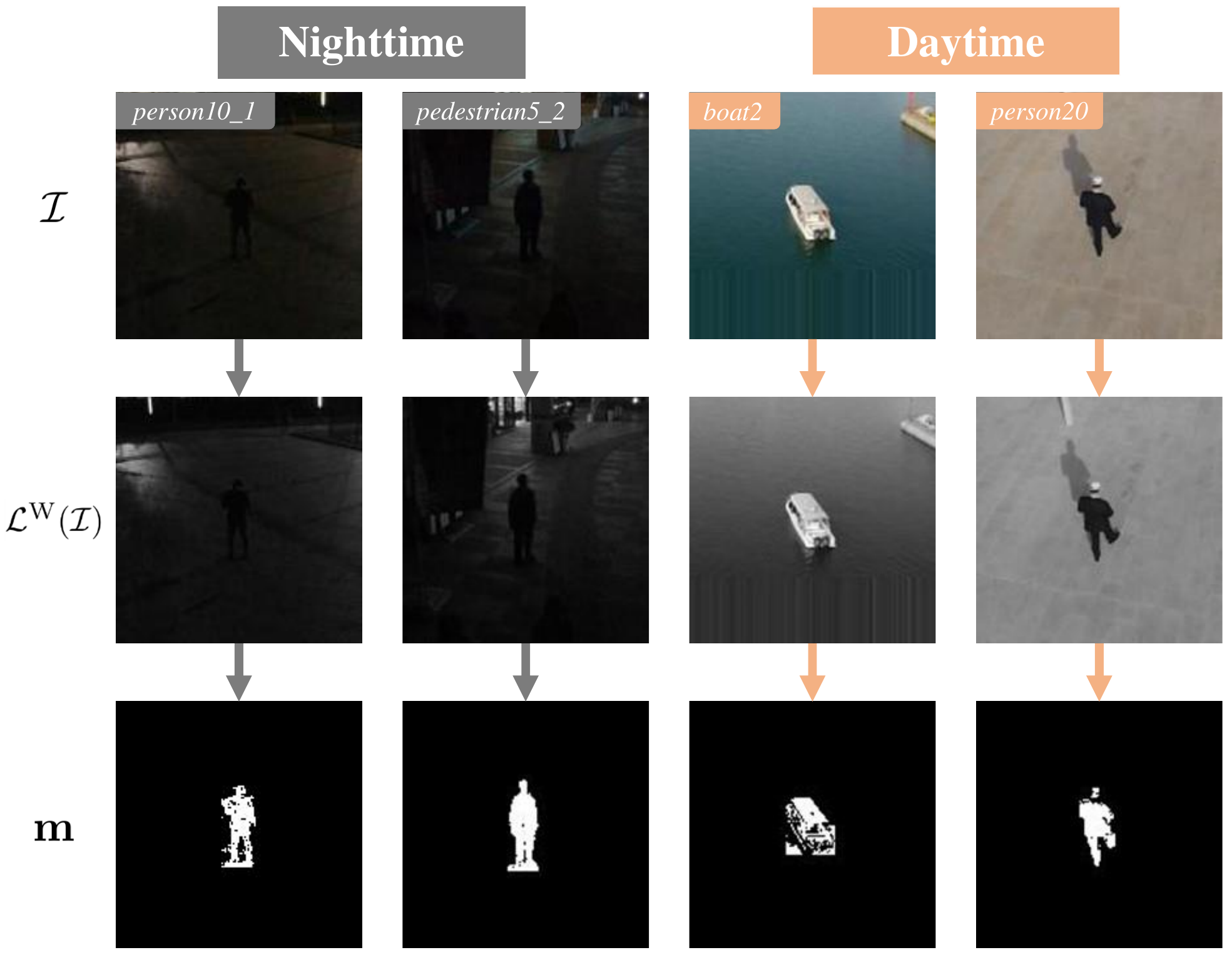}
	\caption{Visualization of mask generation in both nighttime and daytime. From top to bottom, the images denote original patch, illumination map, and generated mask. The sequences \textit{person10\_1}, \textit{pedestrian5\_2} are from newly constructed UAVDark135, and \textit{boat2}, \textit{person20} are from UAV123@10fps \cite{Mueller2016ECCV}. It's clear that in both conditions, the proposed method can obtain valid mask with vivid object contour. }
	\label{fig:mask}
\end{figure}

Ultimately, the expected mask is obtained by $\mathbf{m}=\mathbf{m}_r\odot\mathbf{P}$,
where $\odot$ denotes element-wise product. $\mathbf{P}\in\mathbb{R}^{w\times h}$ is the cropping matrix, which extracts the value of the target-size area in the middle of the raw mask $\mathbf{m}_r$, and set the value of the remaining area to 0 to shield the interference of similar brightness objects in the background. Fig.~\ref{fig:mask} displays some representative examples of mask generation in all-day conditions.

Thus, according to the outcome in Eq.~(\ref{eqn:3}), for $S(\mathcal{I})=1$, ADTrack uses Eq.~(\ref{eqn:5}) to obtain enhanced image $\mathcal{I}_e$, and for $S(\mathcal{I})=0$, original image $\mathcal{I}$ is utilized. Both daytime sequences and nighttime sequences can adopt Eq.~(\ref{eqn:6}) for mask generation. 

\subsection{Filter Training}\label{train}

\subsubsection{Review of BACF}
Due to both its robustness and efficiency, this work adopts background-aware correlation filter (BACF) \cite{Galoogahi2017ICCV} as baseline. The BACF tracker achieves its satisfying performance mainly by virtue of the introduction of the cropping matrix $\mathbf{P}$, which expands the training samples hugely without inletting much boundary effect. The training regression equation of the BACF tracker can be expressed as:
\begin{equation}\label{eqn:8}
\mathcal{E}(\mathbf{w})=\frac{1}{2}\sum_{j=1}^{T}\left\|\sum_{c=1}^{D}\mathbf{w}^{c\top}\mathbf{P}\mathbf{C}^j\mathbf{x}^c-\mathbf{y}(j)\right\|_2^2+\frac{\lambda}{2}\sum_{c=1}^{D}\left\|\mathbf{w}^c\right\|_2^2~,
\end{equation}where $\mathbf{w}^c\in\mathbb{R}^N (c=1,2,\cdots,D)$ is the filter in the $c$-th channel obtained in current frame and $\mathbf{w}=[\mathbf{w}^1,\mathbf{w}^2,\cdots,\mathbf{w}^D]$ denotes the whole filter. $\mathbf{x}^c\in\mathbb{R}^T$ is the $c$-th channel of extracted feature map and $\mathbf{y}(j)$ denotes the $j$-th element in the expected Gaussian-shape regression label $\mathbf{y}\in\mathbb{R}^T$. Cropping matrix $\mathbf{P}\in\mathbb{R}^{N\times T}$ aims at cropping the center region of samples $\mathbf{x}^c$ for training and cyclic shift matrix $\mathbf{C}^j\in\mathbb{R}^{T\times T}$ is the same in \cite{Henriques2015TPAMI}, which is employed to obtain cyclic samples. $\lambda$ is the regularization term parameter.

\Remark Since in Eq.~(\ref{eqn:8}), $T$ and $N$ meet $T>>N$, the filter $\mathbf{w}$ learns far more samples, the negative samples in particular, than in other CF-based trackers. Such strategy makes the filter aware of the background information, resulting in its better discriminative ability.

\subsubsection{Proposed ADTrack}
Apart from BACF~\cite{Galoogahi2017ICCV}, which trains single filter $\mathbf{w}$ with both negative and positive target-size samples, ADTrack trains dual filters $\mathbf{w}_g$ and $\mathbf{w}_{o}$, which learns context and target information separately. Besides, a constraint term is added into the overall objective to promise more robust tracking on-the-fly. The proposed regression objective can be written as:
\vspace{-0.05cm}
\begin{equation}\label{eqn:9}
\begin{split}
\mathcal{E}(\mathbf{w}_g,\mathbf{w}_{o})=&\sum_{k}(\frac{1}{2}\left\|\sum_{c=1}^{D}\mathbf{P}^{ \top}\mathbf{w}_k^c\star\mathbf{x}_k^c-\mathbf{y}\right\|_2^2
\! \! +\frac{\lambda}{2}\sum_{c=1}^{D}\left\|\mathbf{w}_k^c\right\|_2^2)\\
&+\frac{\mu}{2}\sum_{c=1}^{D}\left\|\mathbf{w}_g^c-\mathbf{w}_{o}^c\right\|_2^2~,k\in\{g,o\}~,
\end{split}
\end{equation}
where $\star$ denotes circular correlation operator, which implicitly executes sample augmentation by circular shift. Differently, $\mathbf{x}_g$ denotes the context feature map, while $\mathbf{x}_{o}$ indicates the target region feature map, which is generated using the mask $\mathbf{m}$, \textit{i.e.}, $\mathbf{x}_{o}=\mathbf{m}\odot\mathbf{x}_g$.
The second and fourth term in Eq.~(\ref{eqn:9}) serve as the regularization term to prevent overfitting of the filters. The last term can be considered as the constraint term, where $\mathbf{w}_g$ and $\mathbf{w}_{o}$ bind each other during training. In this case, the discriminative ability of both filters will be more robust. $\mu$ is a parameter used to control the impact of the constraint term.

\Remark In order to maintain historic appearance information of object, this work follows a conventional fashion in \cite{Galoogahi2017ICCV} for adaptive feature updates using linear interpolation strategy with a fixed learning rate $\eta$ as:
\vspace{-0.05cm}
\begin{equation}\label{eqn:up}
\begin{split}
\mathbf{x}^f_{k,\mathrm{model}}=\mathbf{x}^{f-1}_{k,\mathrm{model}}+\eta\mathbf{x}^f_{k}~,k\in\{g,o\}~,
\end{split}
\end{equation}
where $\mathbf{x}^f_{k,\mathrm{model}}$ denotes the training sample in the $f$-th frame, which is utilized to train dual filters in Eq.~(\ref{eqn:9}).

\subsubsection{Optimization}
Assume that $\mathbf{w}_{o}$ is given, ADTrack firstly finds the optimal solution of $\mathbf{w}_g$. Defining an auxiliary variable $\mathbf{v}$, \textit{i.e.}, $\mathbf{v}=\mathbf{I}_N\otimes\mathbf{P}^{\top}\mathbf{w}_g\in\mathbb{R}^{TD}$,
where $\otimes$ denotes Kronecker product, $\mathbf{I}_N$ an $N$-order identical matrix. Here, $\mathbf{w}_g=[\mathbf{w}^{1\top}_g,\mathbf{w}^{2 \top}_g,\cdots,\mathbf{w}^{D\top}_g]^{\top}\in\mathbb{R}^{ND}$. Then, the augmented Lagrangian form of Eq.~(\ref{eqn:9}) is formulated as:
\begin{equation}\label{eqn:10}
\begin{split}
\mathcal{E}(\mathbf{w}_g,\mathbf{v},\bm{\theta})&=\frac{1}{2}\left\|\mathbf{v}\star\mathbf{x}-\mathbf{y}\right\|^2_2+\frac{\lambda}{2}\left\|\mathbf{w}_g\right\|_2^2\\
&+\frac{\mu}{2}\left\|\mathbf{w}_g-\mathbf{w}_{o}\right\|_2^2+(\mathbf{I}_N\otimes\mathbf{P}^{\top}\mathbf{w}_g-\mathbf{v})^{\top}\bm{\theta}\\
&+\frac{\gamma}{2}\left\|\mathbf{I}_N\otimes\mathbf{P}^{\top}\mathbf{w}_g-\mathbf{v}\right\|^2_2~,
\end{split}
\end{equation}
where $\bm{\theta}=[\bm{\theta}^{1\top},\bm{\theta}^{2\top},\cdots,\bm{\theta}^{D \top}]^{\top}\in\mathbb{R}^{TD}$ is the Lagrangian vector and $\gamma$ denotes a penalty factor. Adopting ADMM \cite{Stephen2010FTML}, Eq.~(\ref{eqn:10}) can be dissected and solved by iteratively solving the following three subproblems:
\begin{equation}\label{eqn:11}
\left\{
\begin{aligned}
\mathbf{w}^{e+1}_g&=\rm{arg}\min_{\mathbf{w}}\Big\{\frac{\lambda}{2}\left\|\mathbf{w}^e_g\right\|_2^2+\frac{\mu}{2}\left\|\mathbf{w}^e_g-\mathbf{w}_{o}\right\|_2^2\\
&+(\mathbf{I}_N\otimes\mathbf{P}^{\top}\mathbf{w}^e_g-\mathbf{v})^{\top}\bm{\theta}+\frac{\gamma}{2}\left\|\mathbf{I}_N\otimes\mathbf{P}^{\top}\mathbf{w}^e_g-\mathbf{v}\right\|^2_2\Big\}\\
\mathbf{v}^{e+1}&=\rm{arg}\min_{\mathbf{v}}\Big\{\frac{1}{2}\left\|\mathbf{v}^e\star\mathbf{x}-\mathbf{y}\right\|^2_2+\\
&(\mathbf{I}_N\otimes\mathbf{P}^{\top}\mathbf{w}_g-\mathbf{v}^e)^{\top}\bm{\theta}
+\frac{\gamma}{2}\left\|\mathbf{I}_N\otimes\mathbf{P}^{\top}\mathbf{w}_g-\mathbf{v}^e\right\|^2_2\Big\}\\
\bm{\theta}^{e+1}&=\bm{\theta}^{e}+\gamma(\mathbf{v}^{e+1}-(\mathbf{FP}^{\top}\otimes\mathbf{I}_{D})\mathbf{w}^{e+1}_g)~,\\
\end{aligned}
\right.
\end{equation}
where the superscript $\cdot^{e}$ indicates $e$-th iteration. Following superscript $'$ represents the optimization objectives.

\noindent\textbf{Subproblem $\mathbf{w}'_g$}:
By setting the partial derivative of the first subproblem in Eq.~(\ref{eqn:11}) with respect to $\mathbf{w}_o$ as zero, we can find the closed-form solution of $\mathbf{w}'_g$, which is expressed as:
\begin{equation} \label{eqn:12}
\mathbf{w}'_g=\frac{\mu\mathbf{w}_{o}+T\bm{\theta}+\gamma T \mathbf{v}}{\lambda+\mu+\gamma T}~.
\end{equation}

\noindent\textbf{Subproblem $\mathbf{v}'$}:
To effectively obtain the closed-form of $\mathbf{v}$, this work firstly turn the second subproblem in Eq.~(\ref{eqn:11}) into Fourier domain using discrete Fourier transform (DFT) as:
\begin{equation}\label{eqn:13}
\begin{split}
\mathbf{v}'=\rm{arg}\min_{\hat{\mathbf{v}}}&\Big\{\frac{1}{2T}\left\|\hat{\mathbf{v}}^{*}\odot\hat{\mathbf{x}}-\hat{\mathbf{y}}\right\|^2_2+\hat{\bm{\theta}}^{\top}(\sqrt{T}\mathbf{I}_N\otimes\mathbf{P}^{\top}\mathbf{F}_N\mathbf{w}_g\\
&-\hat{\mathbf{v}})+\frac{\gamma}{2T}\left\|\sqrt{T}\mathbf{I}_N\otimes\mathbf{P}^{\top}\mathbf{F}_N\mathbf{w}_g-\hat{\mathbf{v}}\right\|^2_2\Big\}~,\\
\end{split}
\end{equation}
where $\hat{\cdot}$ denotes the Fourier form of a variable, \textit{i.e.,} $\hat{\mathbf{x}}=\sqrt{T}\mathbf{F}_T\mathbf{x}$. $\mathbf{F}_T\in\mathbb{C}^{T\times T}$ is the Fourier matrix. Superscript $\cdot^{*}$ indicates the complex conjugate. 

\Remark Since circular correlation in time domain is turned into element-wise product in Fourier domain, separating sample in Eq.~(\ref{eqn:13}) across pixels, \textit{e.g.,} $\mathbf{x}(t)=[\mathbf{x}^1(t),\mathbf{x}^2(t),\cdots,\mathbf{x}^D(t)]\in\mathbb{R}^{T\times D}, (t=1,2,\cdots,T)$, each $\hat{\mathbf{v}}'(t)$ can be solved as:
\begin{equation}\label{eqn:14}
\begin{split}
\hat{\mathbf{v}}'(t)=&\Big(\hat{\mathbf{x}}(t)\hat{\mathbf{x}}(t)^{\top}+T\gamma\mathbf{I}_D\Big)^{-1}\\
&\times\Big(\hat{\mathbf{y}}(t)\hat{\mathbf{x}}(t)-T\hat{\bm{\theta}}(t)+T\gamma\hat{\mathbf{w}}_g(t)\Big)~.
\end{split}
\end{equation}

Sherman-Morrison formula \cite{sherman1950AMS} is applied to avoid the time-consuming matrix inversion operation and Eq.~(\ref{eqn:14}) is turned into: 
\begin{equation}\label{eqn:15}
\begin{split}
\hat{\mathbf{v}}'(t)=\frac{1}{\gamma T}\Big(\hat{\mathbf{y}}(t)\hat{\mathbf{x}}(t)-T\hat{\bm{\theta}}(t)+\gamma T\hat{\mathbf{w}}_g(t)\Big)-\\
\frac{\hat{\mathbf{x}}(t)}{\gamma Tb}\Big(\hat{\mathbf{y}}(t)\hat{\mathbf{s}}_{\mathbf{x}}(t)-T\hat{\mathbf{s}}_{\bm{\theta}}(t)+\gamma T\hat{\mathbf{s}}_{\bm{w}_g}(t)\Big)~,
\end{split}
\end{equation}
where $\hat{\mathbf{s}}_{\mathbf{x}}(t)=\hat{\mathbf{x}}(t)^{\top}\hat{\mathbf{x}},\hat{\mathbf{s}}_{\bm{\theta}}=\hat{\mathbf{x}}(t)^{\top}\hat{\mathbf{\theta}},\hat{\mathbf{s}}_{\bm{w}_g}=\hat{\mathbf{x}}(t)^{\top}(t)\hat{\mathbf{w}}_g $, and $b=\hat{\mathbf{s}}_{\mathbf{x}}(t)+T\gamma$ are scalar.

The positions of $\mathbf{w}_g$ and $\mathbf{w}_{o}$ in Eq.~(\ref{eqn:9}) are equivalent. When an solving iteration of $\mathbf{w}_g$ is completed, then the same ADMM iteration operation is performed to obtain the optimized solution of $\mathbf{w}_{o}$. 

\subsection{Target Detection}\label{dete}
Given the expected filter $\mathbf{w}^f_g$ and $\mathbf{w}^f_{o}$ in the $f$-th frame, the response map $\mathbf{R}$ regarding the detection samples $\mathbf{z}^{f+1}$ in the $(f+1)$-th frame can be obtained by:
\begin{equation}
\label{eqn:16}
\begin{split}
\mathbf{R}=\mathcal{F}^{-1}\sum_{c=1}^{D}\big(\hat{\mathbf{w}}^{f,c*}_g\odot\hat{\mathbf{z}}^{f+1,c}_g+\psi\hat{\mathbf{w}}^{f,c*}_{o}\odot\hat{\mathbf{z}}^{f+1,c}_{o}\big)~,
\end{split}
\end{equation}
where $\mathcal{F}^{-1}$ means inverse discrete Fourier transform. $\mathbf{z}_g^{f+1,c}$ denotes the $c$-th channel of resized search region samples extracted in the $(f+1)$-th frame, and $\mathbf{z}_{o}^{f+1,c}$ is the $c$-th channel of the masked samples similar to $\mathbf{x}_{o}$. $\psi$ is a weight parameter that controls the impact response map generated by context filter and object filter. Finally, the object location in the $(f+1)$-th frame can be estimated at the peak of response map $\mathbf{R}$.

The holonomic pipeline pseudo code of ADTrack is summarized in Algorithm~\ref{alg:tracker}.

\normalem
\begin{algorithm}[!t]
	\caption{ADTrack tracker}
	\label{alg:tracker}
	
	\KwIn {A video sequence of $F$ frames.\\
		\hspace{0.95cm} Position ($\mathbf{p}^1$) and size ($\mathbf{s}^1$) of the tracked object \\
		\hspace{0.95cm} in the first frame $\mathcal{I}^1$.}
	\KwOut {Estimated position ($\mathbf{p}^f$) and size ($\mathbf{s}^f$) of the object in all upcoming frames.}
	Construct the Gaussian label function $\mathbf{y}$.\\
	\For{frame number $f=1$ to end}{ 
		\eIf{$f=1$}{
			Calclate log-average illuminance $\tilde{\mathcal{L}}^{\mathrm{W}}(\mathcal{I}^1)$ and dark identifier $S(\mathcal{I}^1)$ and adjust the mode of the tracker (Sec.~\ref{IA}).\\
			Crop the training patch from $\mathcal{I}^1$ with $\mathbf{p}^1$ and $\mathbf{sc} \times \mathbf{s}^1$, where $\mathbf{sc}$ is a predefined scale factor.\\
			\If{$S(\mathcal{I}^1)==1$}{
				Do image enhancing to obtain enhanced patch (Sec.~\ref{pre}).}
			Obtain target-aware mask $\mathbf{m}$ (Sec.~\ref{pre}).\\
			Extract context features $\mathbf{x}^{1}_{g}$ and target features $\mathbf{x}^{1}_{o}$ of the obtained patch.\\
			Update the appearance model $\mathbf{x}^1_{k,\mathrm{model}} = \mathbf{x}^1_{k}~,k\in\{g,o\}$ for filter training.\\
			Learn context and target filter $\mathbf{w}_k~,k\in\{g,o\}$ (Sec.~\ref{train}).\\
		}
		{Crop the search patch from $\mathcal{I}^f$ with $\mathbf{p}^{f-1}$ and $\mathbf{sc} \times \mathbf{s}^{f-1}$.\\
			\If{$S(\mathcal{I}^1)==1$}{
				Do image enhancing to obtain enhanced patch (Sec.~\ref{pre}).}
			Obtain target-aware mask $\mathbf{m}$ (Sec.~\ref{pre}).\\
			Extract context and target search features $\mathbf{z}^{f}_{k}~,k\in\{g,o\}$ of the obtained patch.\\
			Generate the fused response map $\mathbf{R}$ (Sec.~\ref{dete}).\\
			Estimate $\mathbf{p}^f$ and $\mathbf{s}^f$.\\
			Crop the training patch from $\mathcal{I}^f$ with $\mathbf{p}^f$ and $\mathbf{sc} \times \mathbf{s}^f$.\\
			\If{$S(\mathcal{I}^1)==1$}{
				Do image enhancing to obtain enhanced patch (Sec.~\ref{pre}).}
			Obtain target-aware mask $\mathbf{m}$ (Sec.~\ref{pre}).\\
			Extract context features $\mathbf{x}^{f}_{g}$ and target features $\mathbf{x}^{f}_{o}$ of the obtained patch.\\
			Update the appearance model using Eq.~(\ref{eqn:up}) for dual filter training.\\
			Learn context and target filter $\mathbf{w}_k~,k\in\{g,o\}$ (Sec.~\ref{train}).\\}
	}
\end{algorithm} 

\section{UAVDark135 Tracking Benchmark}\label{Sec:4}

\begin{table*}[!t]
	\centering
	\setlength{\tabcolsep}{2mm}
	\fontsize{8}{6}\selectfont
	\begin{threeparttable}
		\caption{Numbers of sequences, minimum, maximum, mean frames in each sequence, and total frames in 6 benchmarks, \textit{i.e.}, newly constructed UAVDark135, UAV123, UAV123@10fps \cite{Mueller2016ECCV}, DTB70 \cite{Li2017AAAI}, UAVDT \cite{du2018ECCV}, and VisDrone2019-SOT \cite{du2019ICCVW}. \textbf{\textcolor[rgb]{ 1,  0,  0}{Red}}, \textbf{\textcolor[rgb]{ 0,  1,  0}{green}}, and \textbf{\textcolor[rgb]{ 0,  0,  1}{blue}} denotes the first, second and third place respectively.}
		\vspace{0.08cm}
		\begin{tabular}{cccccc|ccc|ccc|ccc|ccc|cccc}
			\toprule[1.5pt]
			
			\multicolumn{3}{c}{Benchmark}&\multicolumn{3}{c}{\textbf{UAVDark135}}&\multicolumn{3}{c}{UAV123}&\multicolumn{3}{c}{UAV123@10fps}&\multicolumn{3}{c}{DTB70}&\multicolumn{3}{c}{UAVDT}&\multicolumn{3}{c}{VisDrone2019-SOT}\cr
			\midrule
			
			\multicolumn{3}{c}{Sequences}&\multicolumn{3}{c}{\textbf{\textcolor{red}{135}}}&\multicolumn{3}{c}{\textbf{\textcolor{blue}{123}}}&\multicolumn{3}{c}{\textbf{\textcolor{blue}{123}}}&\multicolumn{3}{c}{70}&\multicolumn{3}{c}{50}&\multicolumn{3}{c}{\textbf{\textcolor{green}{132}}}\cr
			\midrule
			
			\multicolumn{3}{c}{Each sequence}&\multirow{2}{*}{\textbf{\textcolor{red}{216}}}&\multirow{2}{*}{\textbf{\textcolor{red}{4571}}}&\multirow{2}{*}{\textbf{\textcolor{red}{929}}}&\multirow{2}{*}{\textbf{\textcolor{green}{109}}}&\multirow{2}{*}{\textbf{\textcolor{green}{3085}}}&\multirow{2}{*}{\textbf{\textcolor{green}{915}}}&\multirow{2}{*}{37}&\multirow{2}{*}{1029}&\multirow{2}{*}{306}&\multirow{2}{*}{68}&\multirow{2}{*}{699}&\multirow{2}{*}{225}&\multirow{2}{*}{82}&\multirow{2}{*}{2969}&\multirow{2}{*}{742}&\multirow{2}{*}{\textbf{\textcolor{blue}{90}}}&\multirow{2}{*}{\textbf{\textcolor{blue}{2970}}}&\multirow{2}{*}{\textbf{\textcolor{blue}{833}}}\cr
			\cmidrule(lr){1-3} 
			Min&Max&Mean& & & & & & & & & & & & & & & & & & \cr
			\midrule
			
			\multicolumn{3}{c}{Total frames}&\multicolumn{3}{c}{\textbf{\textcolor{red}{125466}}}&\multicolumn{3}{c}{\textbf{\textcolor{green}{112578}}}&\multicolumn{3}{c}{37607}&\multicolumn{3}{c}{15777}&\multicolumn{3}{c}{37084}&\multicolumn{3}{c}{\textbf{\textcolor{blue}{109909}}}\cr
			
			\bottomrule[1.5pt]
			\label{tab:benchmarks}
			\vspace{-0.4cm}
		\end{tabular}
	\end{threeparttable}
\end{table*}

\begin{table}[!t]
	\centering
	\fontsize{8}{8}
	\selectfont
	\caption{Detailed explanation of the attributes in newly built UAVDark135, which are commonly confronted in UAV tracking.}
	\setlength{\tabcolsep}{1mm}
	\begin{tabular}{@{}c|c@{}}
		\toprule[1.5pt]
		Attributes&Explanation \\
		\midrule
		VC & \tabincell{c}{\textit{Viewpoint Change}: In the sequence, different aspects, \textit{e.g.},\\ front, side, and top aspect, of the \\tracking object are captured and involved.}\\
		\midrule
		FM& \tabincell{c}{\textit{Fast Motion}: There exist two continuous frames, where the\\ center locations of the tracking object change\\ more than 20 pixels.} \\
		\midrule
		LR &\tabincell{c}{\textit{Low Resolution}: There exist frames, where the\\ tracking object is small, whose total resolution is\\ fewer than 20 pixels.}\\
		\midrule
		OCC &\tabincell{c}{\textit{Occlusion}: There exist frames, where the\\ tracking object is partially or\\ fully occluded by obstacles.}\\
		\midrule
		IV &\tabincell{c}{\textit{Illumination Variation}: In the sequence, the\\ tracking object undergoes various light conditions.}\\
		\bottomrule[1.5pt]			
	\end{tabular}\label{tab:att}
\end{table}

\subsection{Platform and Statistics}
Standing as the first UAV dark tracking benchmark, the UAVDark135 contains totally 135 sequences captured by a standard UAV\footnote{This work utilized Parrot Bebop 2 drone as shooting platform. More detailed information can be found at \url{https://support.parrot.com/us/support/products/parrot-bebop-2}.} at night. The benchmark includes various tracking scenes, \textit{e.g.}, crossings, t-junctions, road, highway, and consists of different kinds of tracked objects like people, boat, bus, car, truck, athletes, house, \textit{etc.} To extent the covered scenes, the benchmark also contains some sequences from YouTube, which were shot on the sea. The total frames, mean frames, maximum frames, and minimum frames of the benchmark are 125466, 929, 4571, and 216 respectively, making it suitable for large-scale evaluation. TABLE~\ref{tab:benchmarks} exhibits main statistics of UAVDark135 against existing UAV tracking benchmarks, \textit{i.e.}, UAV123@10fps, UAV123 \cite{Mueller2016ECCV}, DTB70 \cite{Li2017AAAI}, UAVDT \cite{du2018ECCV}, and VisDrone2019-SOT \cite{du2019ICCVW} (VisDrone2019-SOT testset-challenge is not included). The videos are captured at a frame-rate of 30 frames/s (FPS), with the resolution of 1920$\times$1080.

\Remark Despite the fact that there exist some sequences captured at night in benchmark UAVDT \cite{du2018ECCV} and VisDrone2019-SOT \cite{du2019ICCVW}, it is far from an exhaustive dark tracking evaluation. Besides, the night sequences are actually well-illuminated in \cite{du2018ECCV,du2019ICCVW}, which can not represent more common dark tracking scenes in UAVDark135, where the light conditions are much more hash.

\subsection{Annotation}
The frames in UAVDark135 are all manually annotated, where a sequence is completely processed by the same annotator to ensure consistency. Since in some dark scenes the object is nearly invisible, annotation process is much more strenuous. After the first round, 5 professional annotators carefully checked the results and made revision for several rounds to reduce errors as much as possible in nearly 2 months.

Since the boundary contour of the object is not obvious in the dark, the result boxes of the first annotation fluctuates in continuous image frames. However, the actual motion process of the object should be smooth. In these considerations, we record the original annotation every 5 frames for the sequence with extremely severe vibration, and the results of the remaining frames are obtained by linear interpolation, which is closer to the position and scale variation of the real object.

\begin{figure}[!t]
	\vspace{-0.3cm}
	\centering
	\setlength{\abovecaptionskip}{-0.3cm}
	\includegraphics[width=1\columnwidth]{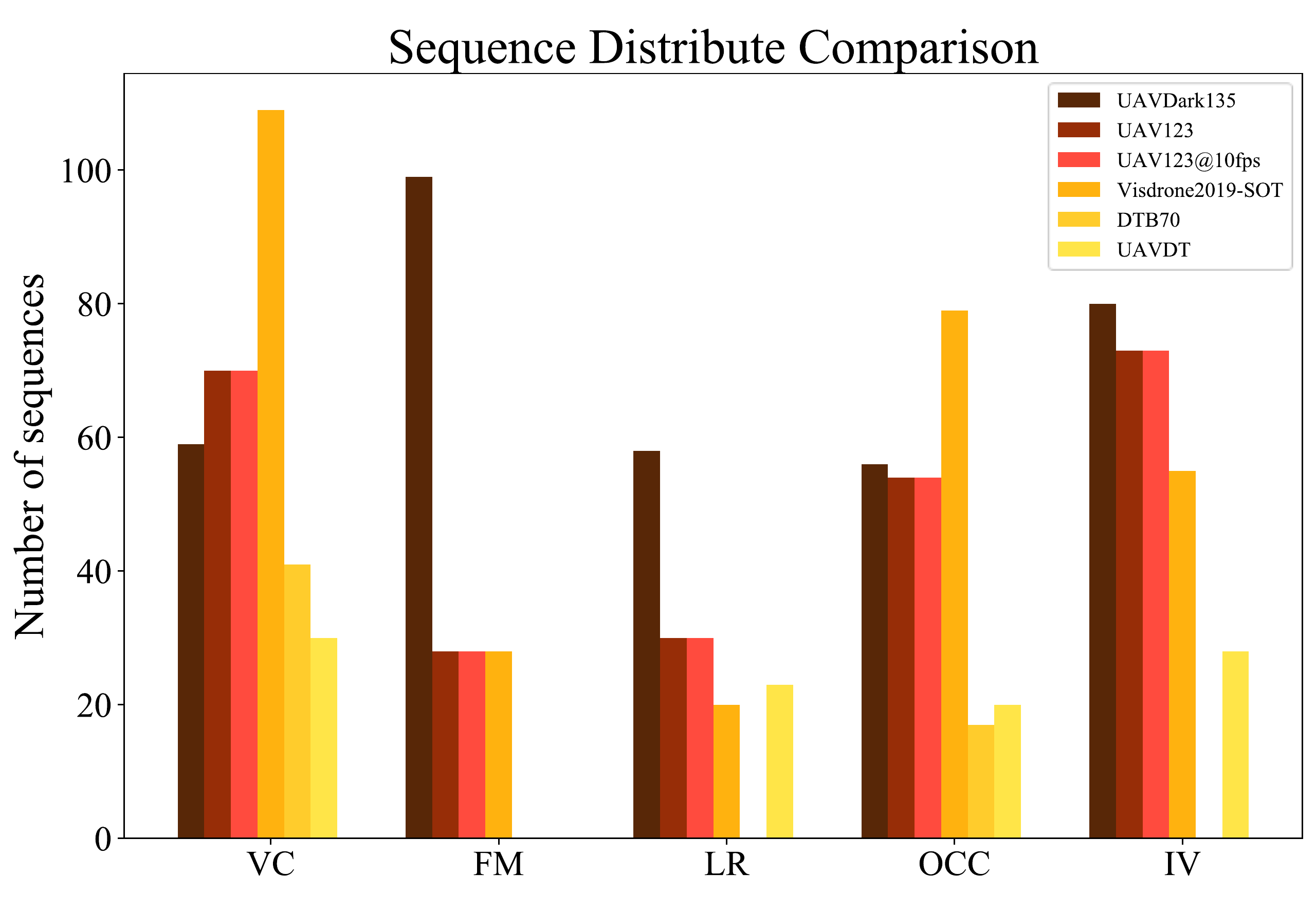}
	\caption{Sequences distribution comparison of 6 UAV tracking benchmarks. The abscissa are 5 attributes, and the ordinate is the numbers of the sequences. The sequences in different benchmarks are marked by different colors, which are explained in the legend. Note that not all benchmarks made contribution to all 5 attributes.
	}
	\label{fig:att}
\end{figure}

\begin{figure*}[!t]
	\begin{center}
		\subfigure{\label{fig:10fps} 
			\begin{minipage}{0.31\textwidth}
				\centering
				\includegraphics[width=1\columnwidth]{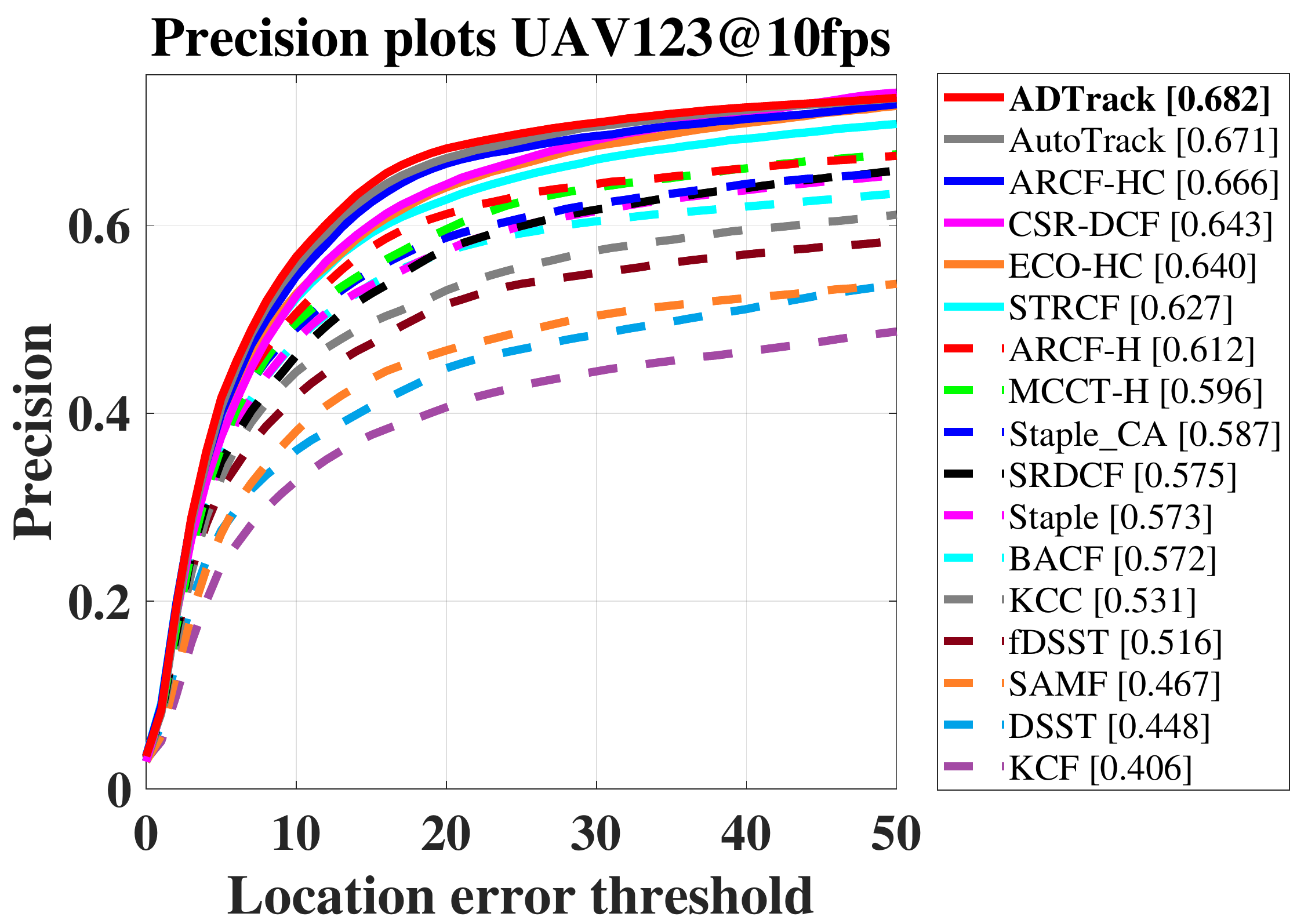}
			\end{minipage}
		}
		\subfigure{\label{fig:DTB70} 
			\begin{minipage}{0.31\textwidth}
				\centering
				\includegraphics[width=1\columnwidth]{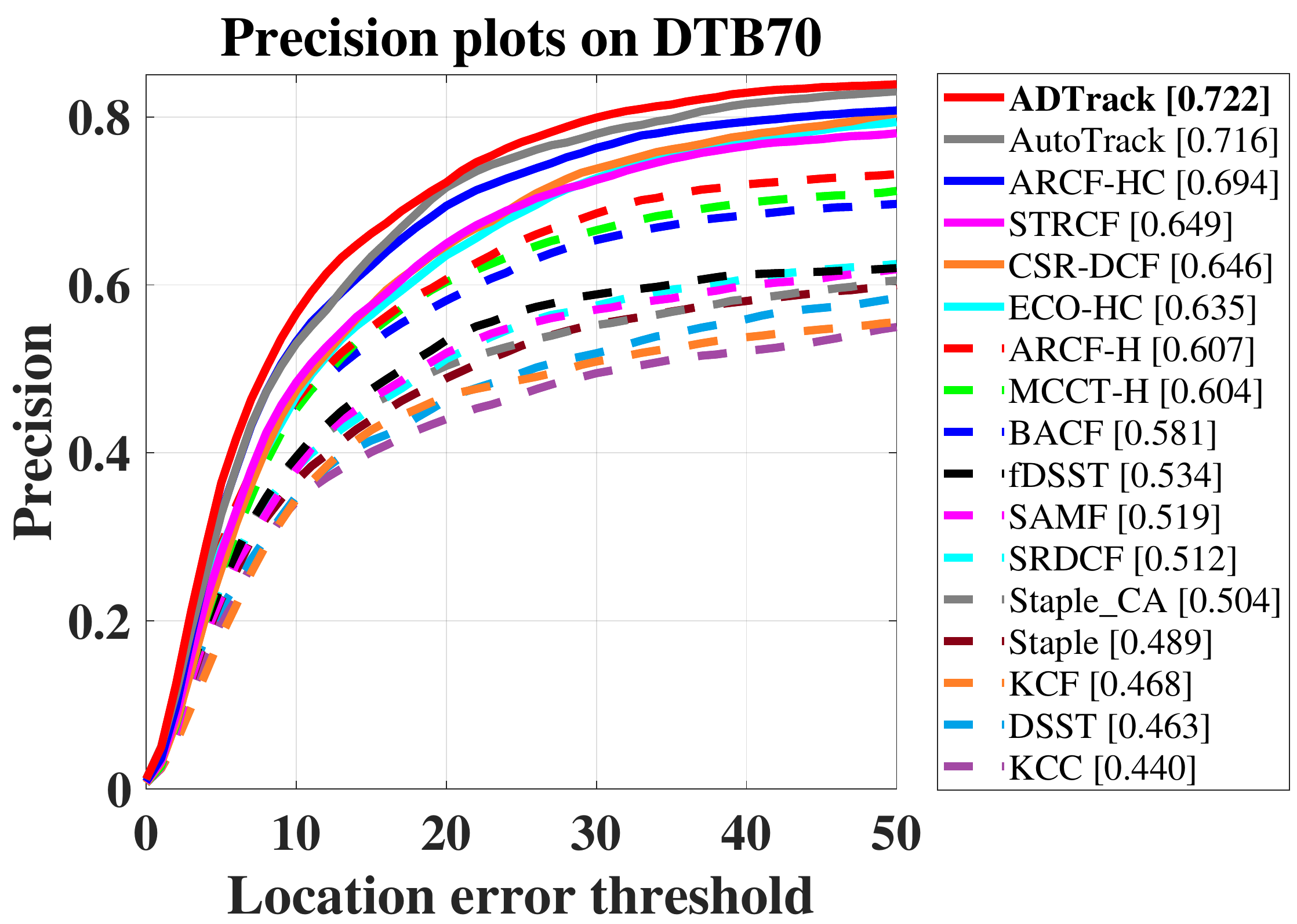}
			\end{minipage}
		}
		\subfigure{\label{fig:UAVDark} 
			\begin{minipage}{0.31\textwidth}
				\centering
				\includegraphics[width=1\columnwidth]{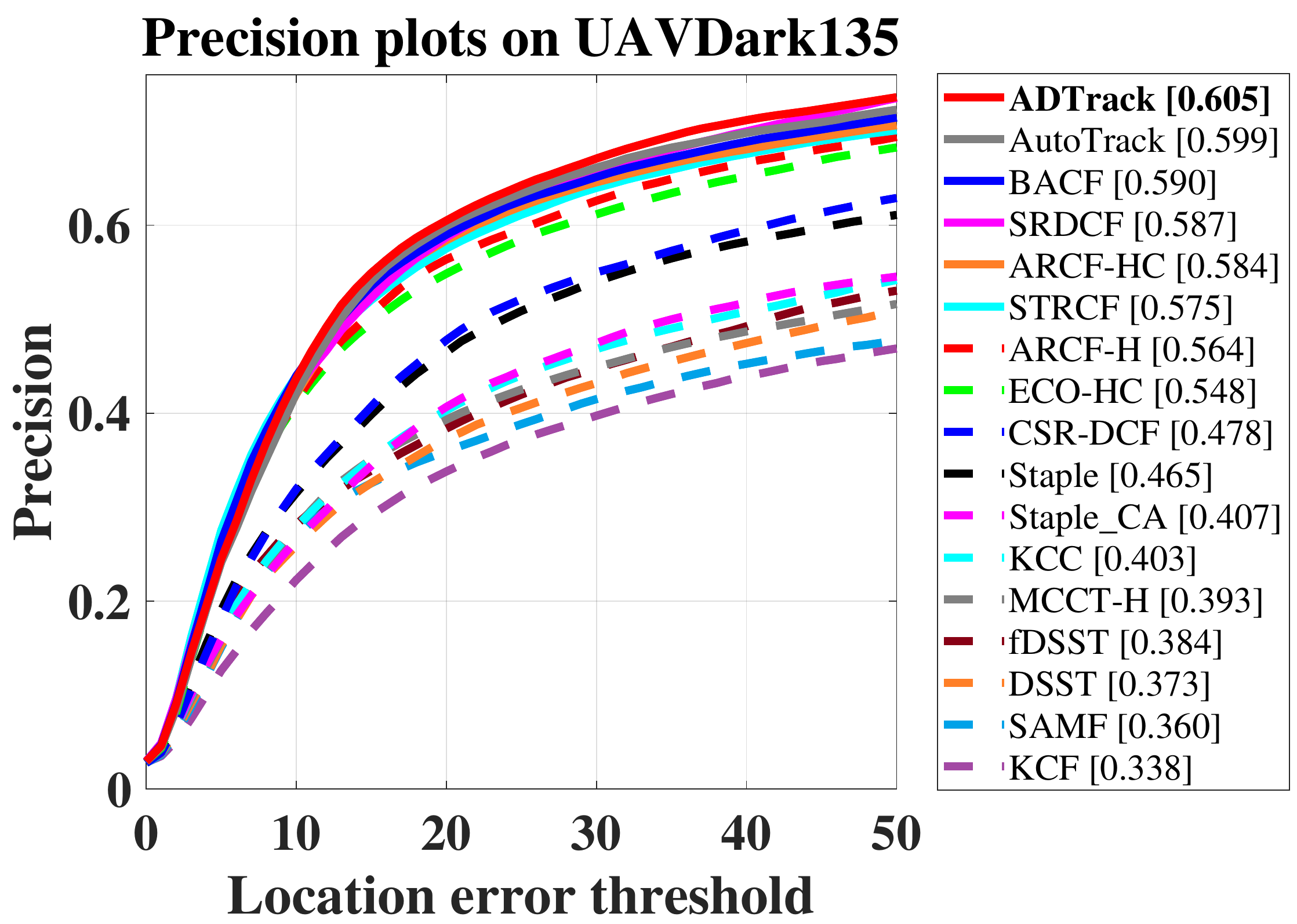}
			\end{minipage}
		}
		
		\subfigure{
			\begin{minipage}{0.31\textwidth}
				\centering
				\includegraphics[width=1\columnwidth]{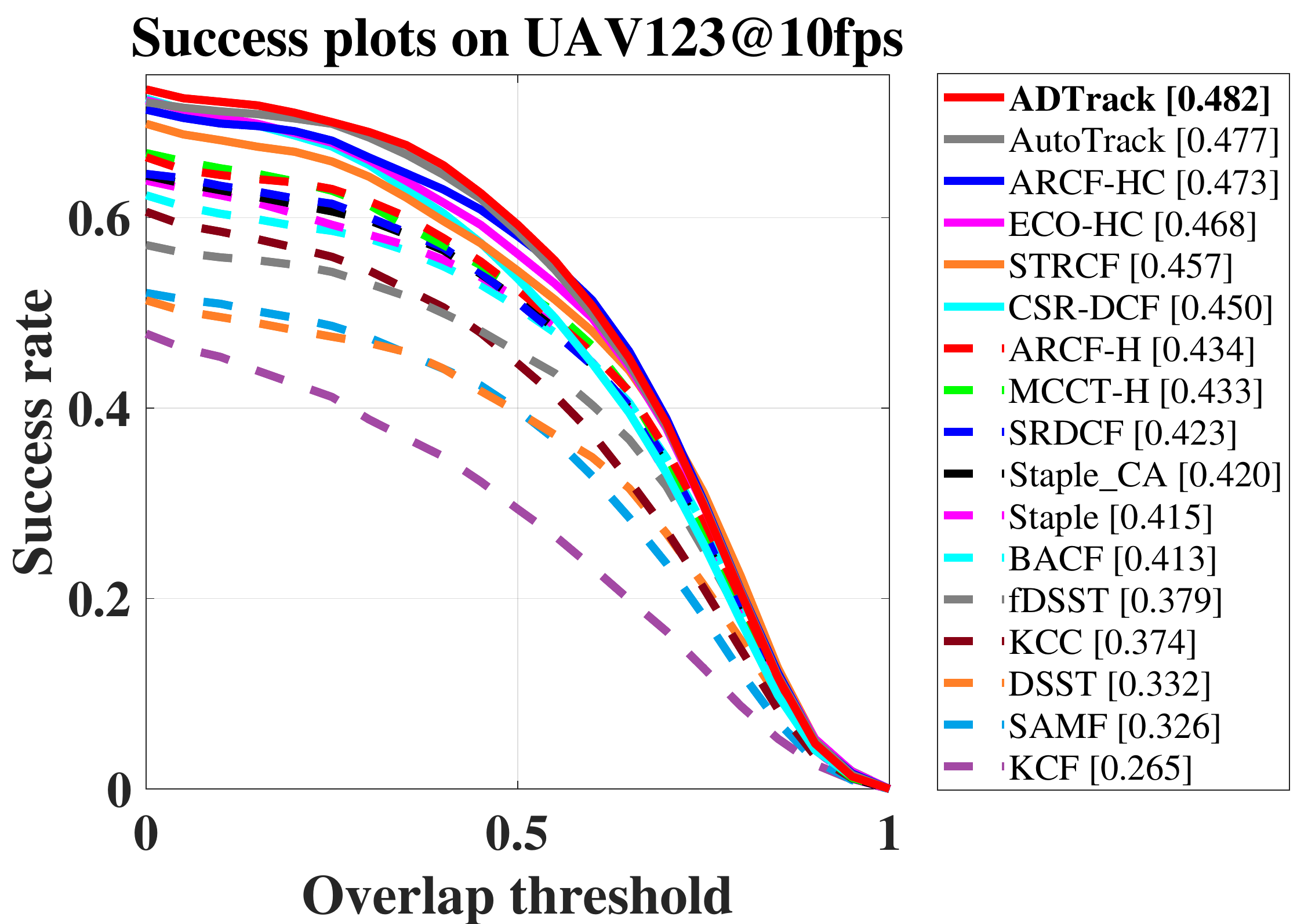}
				(a)
			\end{minipage}
		}
		\subfigure{
			\begin{minipage}{0.31\textwidth}
				\centering
				\includegraphics[width=1\columnwidth]{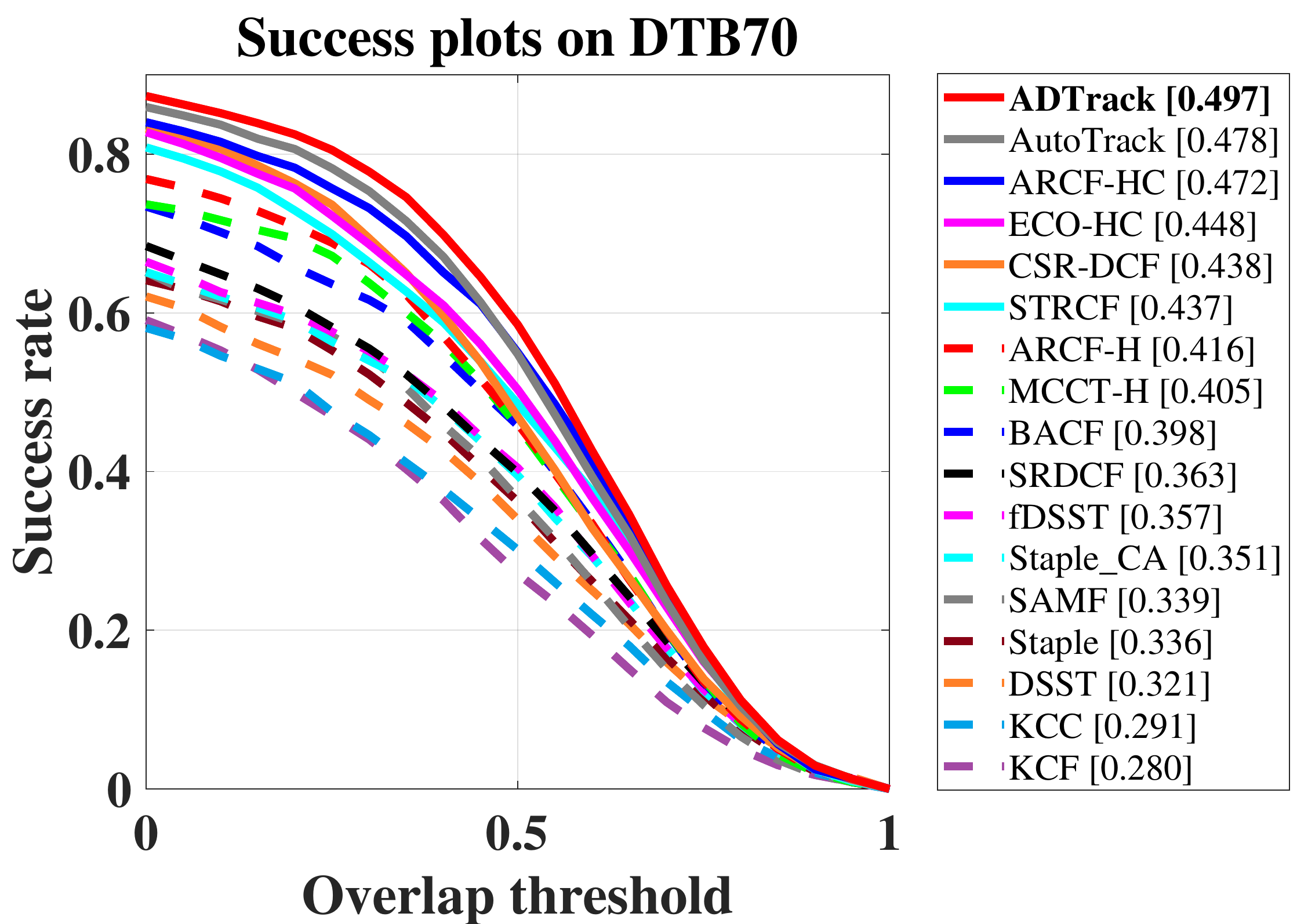}
				(b)
			\end{minipage}
		}
		\subfigure{
			\begin{minipage}{0.31\textwidth}
				\centering
				\includegraphics[width=1\columnwidth]{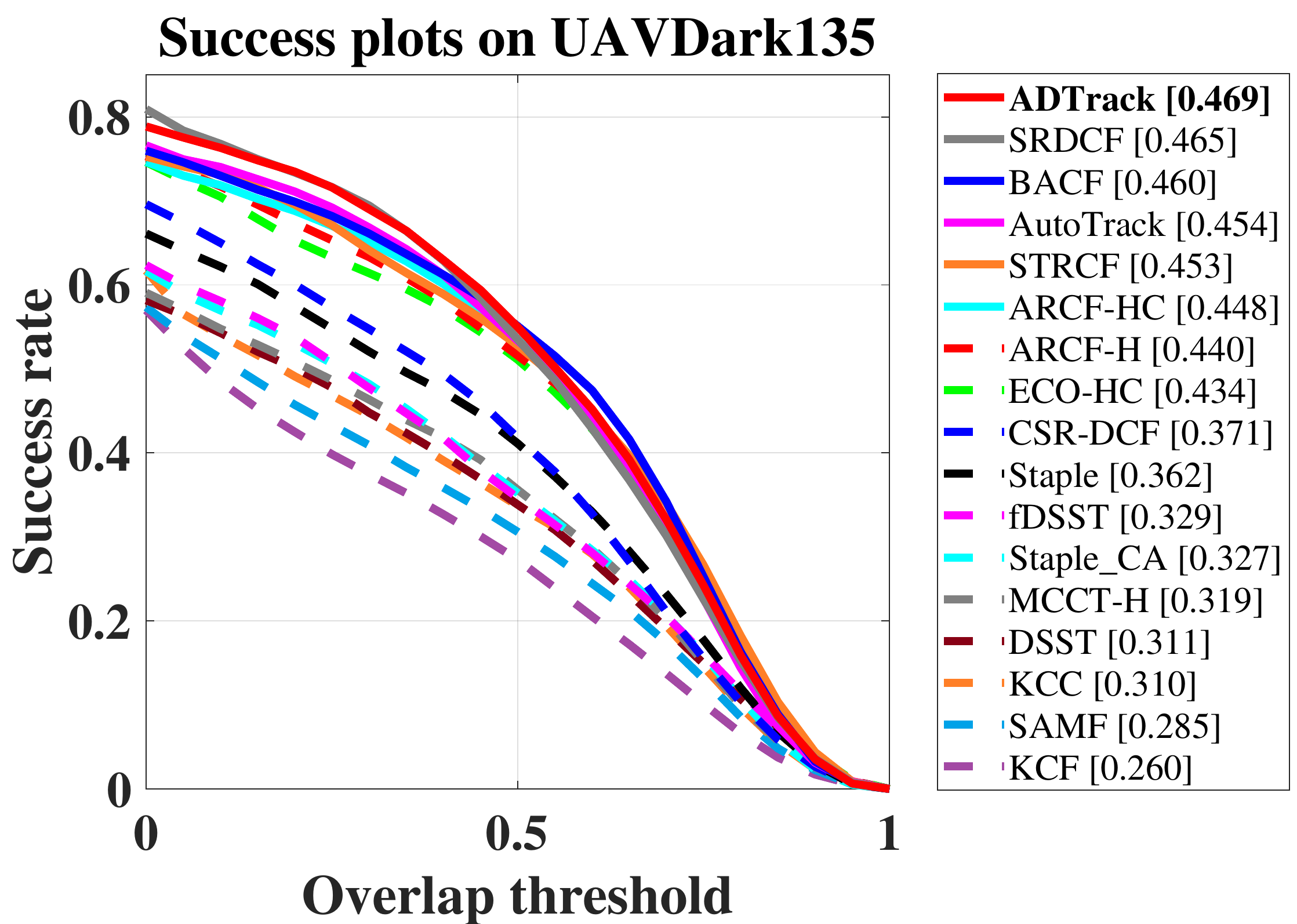}
				(c)
			\end{minipage}
		}
	\end{center}
	\caption{Overall performance of SOTA handcrafted CF-based trackers on UAV123@10fps \cite{Mueller2016ECCV}, DTB70 \cite{Li2017AAAI}, and newly built UAVDark135. The evaluation metric in precision plot is the distance precision (DP) under center location error (CLE) = 20 pixels, and the metric in success rate plot is the area under curve (AUC). Clearly ADTrack maintains its robustness in all 3 benchmarks by virtue of its dual regression.}
	\label{fig:overall}
	\vspace{-0.3cm}
\end{figure*}

\subsection{Attributes}
To better evaluate the trackers' abilities under special challenges, UAVDark135 also provides 5 commonly encountered challenge attributes in UAV tracking, following our prior work \cite{Fu2020GRSM}, \textit{i.e.}, viewpoint change (VC), fast motion (FM), low resolution (LR), occlusion (OCC), and illumination variation (IV). TABLE~\ref{tab:att} elaborately explains the criterion for each attribute. Additionally, Fig.~\ref{fig:att} displays sequences distribute comparison of 6 UAV tracking benchmarks. Clearly, UAVDark135 distributes both evenly and considerably in the five attributes.

\section{Experiment and Evaluation}\label{Sec:5}

\begin{table*}[!t]
	\centering
	\setlength{\tabcolsep}{.6mm}
	\fontsize{6}{8}\selectfont
	\caption{Average results by all 328 sequences of handcrafted trackers on benchmarks UAV123@10fps \cite{Mueller2016ECCV}, DTB70 \cite{Li2017AAAI}, and newly constructed UAVDark135. \textbf{\textcolor[rgb]{ 1,  0,  0}{Red}}, \textbf{\textcolor[rgb]{ 0,  1,  0}{green}}, and \textbf{\textcolor[rgb]{ 0,  0,  1}{blue}} denotes the first, second and third place respectively. Here, the abilities of the trackers under all-day conditions are evaluated.}
	\begin{tabular}{cccccccccccccccccc}
		\toprule[1.5pt]
		Tracker & \multicolumn{1}{c}{\bf{ADTrack}} & \multicolumn{1}{c}{AutoTrack} & \multicolumn{1}{c}{ARCF-HC} & \multicolumn{1}{c}{ARCF-H} & \multicolumn{1}{c}{MCCT-H} & \multicolumn{1}{c}{STRCF} & \multicolumn{1}{c}{KCC} & \multicolumn{1}{c}{fDSST} & \multicolumn{1}{c}{DSST} & \multicolumn{1}{c}{BACF} & \multicolumn{1}{c}{CSR-DCF} & \multicolumn{1}{c}{ECO-HC} & \multicolumn{1}{c}{Staple\_CA} & \multicolumn{1}{c}{Staple} & \multicolumn{1}{c}{KCF} & \multicolumn{1}{c}{SRDCF} & \multicolumn{1}{c}{SAMF} \\
		\midrule
		Venue & \multicolumn{1}{c}{\bf{Ours}} & \multicolumn{1}{c}{'20CVPR} & \multicolumn{1}{c}{'19ICCV} & \multicolumn{1}{c}{'19ICCV} & \multicolumn{1}{c}{'18CVPR} & \multicolumn{1}{c}{'18CVPR} & \multicolumn{1}{c}{'18AAAI} & \multicolumn{1}{c}{'17TPAMI} & \multicolumn{1}{c}{'17TPAMI} & \multicolumn{1}{c}{'17ICCV} & \multicolumn{1}{c}{'17CVPR} & \multicolumn{1}{c}{'17CVPR} & \multicolumn{1}{c}{'17CVPR} & \multicolumn{1}{c}{'16CVPR} & \multicolumn{1}{c}{'15TPAMI} & \multicolumn{1}{c}{'15ICCV} & \multicolumn{1}{c}{'14ECCV} \\
		DP    & \textcolor[rgb]{ 1,  0,  0}{\textbf{0.659}} & \textcolor[rgb]{ 0,  1,  0}{\textbf{0.651}} & \textcolor[rgb]{ 0,  0,  1}{\textbf{0.638}} & 0.591 & 0.514 & 0.611 & 0.459 & 0.465 & 0.420 & 0.581 & 0.576 & 0.601 & 0.495 & 0.510 & 0.391 & 0.566 & 0.434 \\
		AUC   & \textcolor[rgb]{ 1,  0,  0}{\textbf{0.480}} & \textcolor[rgb]{ 0,  1,  0}{\textbf{0.468}} & \textcolor[rgb]{ 0,  0,  1}{\textbf{0.462}} & 0.433 & 0.380 & 0.451 & 0.330 & 0.354 & 0.321 & 0.429 & 0.415 & 0.449 & 0.367 & 0.377 & 0.266 & 0.427 & 0.312 \\
		FPS   & 31.621 & 45.485 & 22.585 & 32.320 & 44.858 & 20.438 & 29.393 & \textcolor[rgb]{ 0,  1,  0}{\textbf{122.976}} & 58.113 & 33.911 & 9.274 & 53.571 & 46.829 & \textcolor[rgb]{ 0,  0,  1}{\textbf{81.216}} & \textcolor[rgb]{ 1,  0,  0}{\textbf{374.912}} & 8.583 & 7.518 \\
		
		\bottomrule[1.5pt]
	\end{tabular}%
	\label{tab:AVE}%
	\vspace{-0.4cm}
\end{table*}%

This section displays exhaustive experimental evaluations, involving night benchmark UAVDark135, daytime benchmark UAV123@10fps \cite{Mueller2016ECCV}, and DTB70 \cite{Li2017AAAI}. In subsection \ref{p1}, implementation details including experiment platform, parameter settings, features, and metrics are introduced. Subsection \ref{p2} gives a comprehensive comparison of the handcrafted CF-based trackers on the benchmarks, where the superiority of proposed ADTrack for all-day UAV tracking is demonstrated. Subsection \ref{p3} presents attribute-based evaluation of the handcrafted CF-based trackers to test their abilities under UAV special challenges. In subsection \ref{p4}, we also invite SOTA deep trackers that utilize convolution neural network (CNN) for dark tracking comparison. Lastly, in subsection \ref{p5}, we extended ablation study and parameter analysis to further demonstrate the validity of different modules in proposed ADTrack.
\subsection{Implementation Details}\label{p1}
\subsubsection{Platform}
The experiments extended in this work were mainly performed on MATLAB R2019a. The main hardware adopted consists of an Intel Core I7-8700K CPU and 32GB RAM.

\subsubsection{Parameters}
To guarantee the fairness and objectivity of the evaluation, the tested trackers from other works have maintained their official initial parameters.

The different parameters of two conditions in ADTrack are as follows: In daytime mode, $\mu$ is set as 280. During detection, weight $\psi$ is set as $0.02$. Translation filter takes learning rate $\eta_t=0.032$ for model update and for scale filter, $\eta_s$ is set as 0.016. In nighttime mode, $\mu$ is set as 200. During detection, weight $\psi$ is set as $0.01$. Translation filter takes learning rate $\eta_t=0.024$ for model update and for scale filter, $\eta_s$ is set as 0.023.

\Remark ADTrack can adapt to the given light condition in its first stage, where the tracking mode mentioned is switched automatically without manually adjusting.

\subsubsection{Features and Scale Estimation}
ADTrack uses handcrafted features for appearance representations, \textit{i.e.}, gray-scale, a fast version of histogram of oriented gradient (fHOG) \cite{Felzenszwalb2010TPAMI}, and color names (CN) \cite{Weijer2006ECCV}. Note that gray-scale and CN features can be valid in ADTrack thanks to low-light enhancement. The cell size for feature extraction is set as $4\times 4$. ADTrack adopts the scale filter proposed by \cite{danelljan2017TPAMI} to perform accurate scale estimation.

\subsubsection{Metrics}
In the experiment evaluation, we mainly use two metrics: distance precision (DP) and area under curve (AUC). DP is based on the distance between the center points of the predicted box and the target ground-truth, and AUC is based on the intersection ratio of the predicted box and the target ground-truth box.

\subsection{Overall Evaluation}\label{p2}
Using merely handcrafted features, most handcrafted CF-based trackers can achieve satisfying running speed, by virtue of their light calculation, while ensuring their robustness under various tracking scenes onboard UAV. This work employs proposed ADTrack and 16 SOTA handcrafted CF-based trackers, \textit{i.e.}, AutoTrack \cite{Li2020CVPR}, KCF \cite{Henriques2015TPAMI}, SAMF \cite{Li2014ECCV}, SRDCF \cite{Danelljan2015ICCV}, STRCF \cite{Li2018CVPR}, BACF \cite{Galoogahi2017ICCV}, DSST \& fDSST \cite{danelljan2017TPAMI}, ECO-HC \cite{Danelljan2017CVPR}, ARCF-HC \& ARCF-H \cite{Huang2019ICCV}, KCC \cite{wang2018AAAI}, MCCT-H \cite{wang2018CVPR}, CSR-DCF \cite{Lukezic2017CVPR}, Staple \cite{bertinetto2016CVPR}, and Staple\_CA \cite{mueller2017CVPR}, for evaluation on tracking benchmarks, UAV123@10fps \cite{Mueller2016ECCV}, DTB70 \cite{Li2017AAAI}, and UAVDark135 to demonstrate the robustness of the proposed ADTrack in all-day UAV tracking comprehensively.

\begin{figure*}[!t]
	\centering
	\includegraphics[width=2\columnwidth]{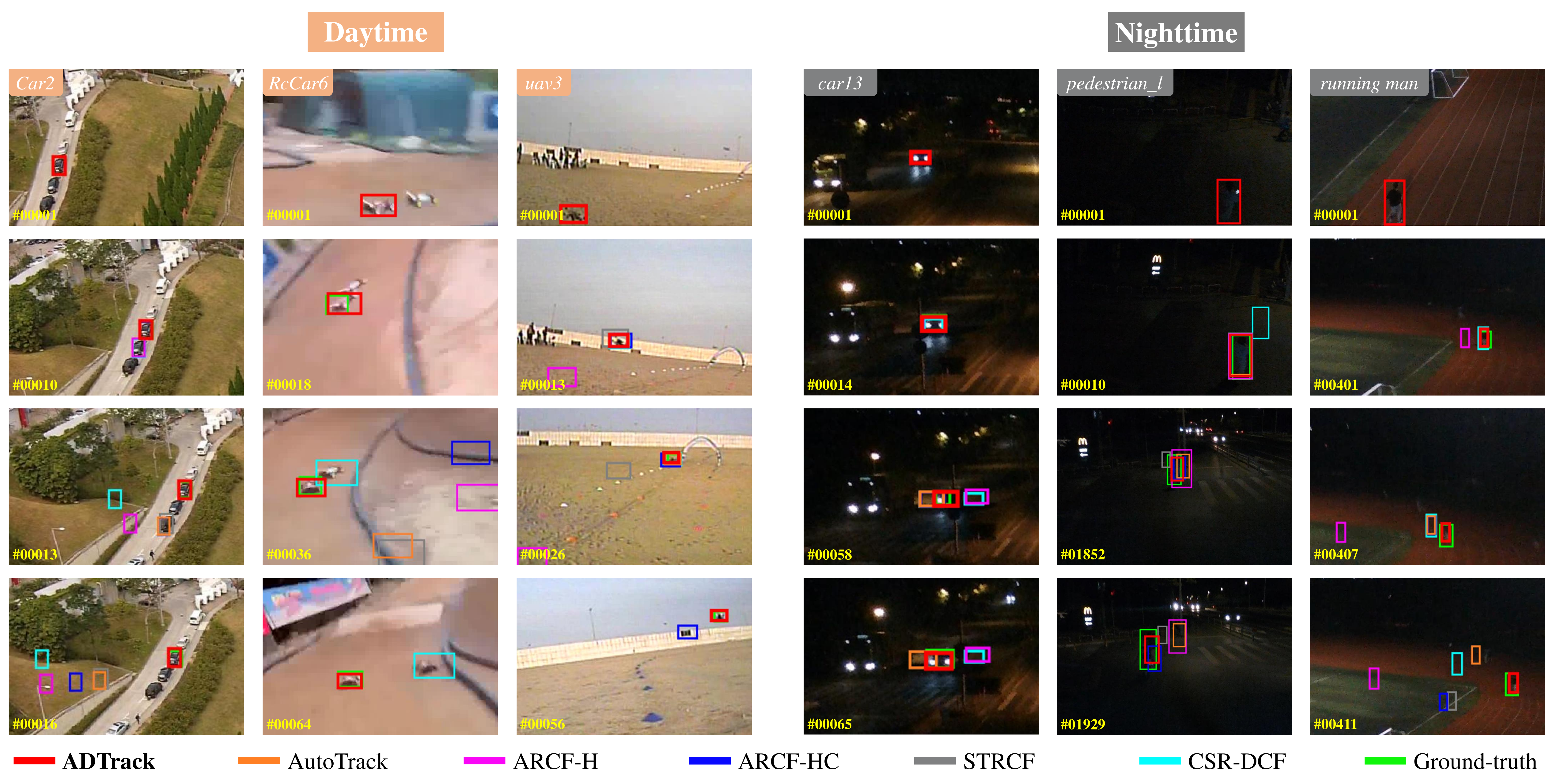}
	\caption{Visualization of some typical tracking scenes in both daytime and nighttime. Sequences \textit{Car2}, \textit{RcCar6}, and \textit{uav3} are from the daytime benchmarks DTB70 and UAV123@10fps. Sequences \textit{car13}, \textit{pedestrian\_l}, and \textit{running\_man} are from the new nighttime benchmark UAVDark135. Clearly, ADTrack favorably maintains its robustness in all-day UAV tracking challenges. The typical sequences were made into video, which can be found at \url{https://youtu.be/cJMUKF4J38A}. }
	\label{fig:vis}
	\vspace{-0.4cm}
\end{figure*}


\subsubsection{Daytime Performance}
In Fig.~\ref{fig:overall}, DP and AUC comparison on benchmarks, respectively, UAV123@10fps \cite{Mueller2016ECCV}, DTB70 \cite{Li2017AAAI} is exhibited in the first 2 columns, where ADTrack ranks first in both metrics. Specifically, in Fig.~\ref{fig:10fps}, ADTrack surpasses the second-best AutoTrack tracker (0.671) \cite{Li2020CVPR} by 1.6\% (0.682). In terms of AUC, ADTrack surpasses its baseline BACF tracker (0.413) \cite{Galoogahi2017ICCV} by over \textbf{19\%} (0.482). Fig.~\ref{fig:DTB70} shows that ADTrack brings its baseline (0.581) up by \textbf{24\%} (0.722) in DP, and exceeds the brilliant AutoTrack tracker (0.478) by nearly \textbf{4\%} (0.497) in AUC. The outstanding results achieved by ADTrack in daytime indicates its strong robustness in real-world UAV tracking scenes by virtue of its dual filter learning.

\subsubsection{Nighttime Performance}
Fig.~\ref{fig:UAVDark} displays the excellent handcrafted CF-based trackers' DPs under newly constructed benchmark UAVDark135, where clearly ADTrack exceeds all other trackers, surpassing the second best tracker (0.599) by over \textbf{1\%} (0.605). Additionally, ADTrack enjoys its satisfying advantages in success rate as well, exceeding the baseline tracker (0.460) by over \textbf{1.9\%} (0.469). In dark scenes like UAVDark135, ADTrack maintains its robustness, providing a favorable choice for UAV all-day tracking.

\begin{table}[!b]
	\centering
	\setlength{\tabcolsep}{1.1mm}
	\fontsize{6}{8}\selectfont
	\caption{Average performance of the handcrafted trackers by UAV special attributes. Obviously, ADTrack maintains its superiority in most challenges.}
	\begin{tabular}{ccccccccccc}
		\toprule[1.5pt]
		\multirow{2}{*}{\diagbox{Tracker}{Metric}}&
		\multicolumn{5}{c}{DP}&
		\multicolumn{5}{c}{AUC} \\
		\cmidrule(lr){2-6} \cmidrule(lr){7-11}    & VC    & FM    & LR    & OCC   & IV    & VC    & FM    & LR    & OCC   & IV \\
		\midrule
		\bf{ADTrack} & \textcolor[rgb]{ 1,  0,  0}{\textbf{0.637}} & \textcolor[rgb]{ 1,  0,  0}{\textbf{0.63}} & \textcolor[rgb]{ 1,  0,  0}{\textbf{0.668}} & \textcolor[rgb]{ 1,  0,  0}{\textbf{0.622}} & \textcolor[rgb]{ 1,  0,  0}{\textbf{0.605}} & \textcolor[rgb]{ 1,  0,  0}{\textbf{0.464}} & \textcolor[rgb]{ 1,  0,  0}{\textbf{0.464}} & \textcolor[rgb]{ 1,  0,  0}{\textbf{0.471}} & \textcolor[rgb]{ 1,  0,  0}{\textbf{0.434}} & \textcolor[rgb]{ 1,  0,  0}{\textbf{0.437}} \\
		AutoTrack & \textcolor[rgb]{ 0,  1,  0}{\textbf{0.622}} & \textcolor[rgb]{ 0,  0,  1}{\textbf{0.588}} & \textcolor[rgb]{ 0,  1,  0}{\textbf{0.651}} & \textcolor[rgb]{ 0,  0,  1}{\textbf{0.598}} & \textcolor[rgb]{ 0,  1,  0}{\textbf{0.599}} & \textcolor[rgb]{ 0,  1,  0}{\textbf{0.448}} & \textcolor[rgb]{ 0,  0,  1}{\textbf{0.433}} & \textcolor[rgb]{ 0,  0,  1}{\textbf{0.455}} & 0.412 & 0.431 \\
		ARCF-HC & \textcolor[rgb]{ 0,  0,  1}{\textbf{0.61}} & \textcolor[rgb]{ 0,  1,  0}{\textbf{0.61}} & \textcolor[rgb]{ 0,  0,  1}{\textbf{0.649}} & 0.595 & \textcolor[rgb]{ 0,  0,  1}{\textbf{0.597}} & \textcolor[rgb]{ 0,  0,  1}{\textbf{0.438}} & \textcolor[rgb]{ 0,  1,  0}{\textbf{0.448}} & \textcolor[rgb]{ 0,  1,  0}{\textbf{0.458}} & \textcolor[rgb]{ 0,  0,  1}{\textbf{0.417}} & \textcolor[rgb]{ 0,  0,  1}{\textbf{0.433}} \\
		ARCF-H & 0.565 & 0.551 & 0.606 & 0.537 & 0.56  & 0.413 & 0.4   & 0.427 & 0.373 & 0.411 \\
		MCCT-H & 0.504 & 0.471 & 0.504 & 0.503 & 0.476 & 0.366 & 0.361 & 0.367 & 0.353 & 0.353 \\
		STRCF & 0.584 & 0.568 & 0.611 & 0.584 & 0.59  & 0.424 & 0.43  & 0.442 & 0.406 & \textcolor[rgb]{ 0,  1,  0}{\textbf{0.437}} \\
		KCC   & 0.425 & 0.451 & 0.493 & 0.427 & 0.459 & 0.309 & 0.329 & 0.348 & 0.297 & 0.326 \\
		fDSST & 0.462 & 0.406 & 0.481 & 0.436 & 0.424 & 0.343 & 0.327 & 0.363 & 0.317 & 0.329 \\
		DSST  & 0.425 & 0.342 & 0.413 & 0.385 & 0.391 & 0.316 & 0.275 & 0.303 & 0.274 & 0.298 \\
		BACF  & 0.55  & 0.554 & 0.582 & 0.517 & 0.537 & 0.41  & 0.411 & 0.414 & 0.371 & 0.402 \\
		CSR-DCF & 0.57  & 0.536 & 0.576 & 0.561 & 0.54  & 0.399 & 0.383 & 0.405 & 0.387 & 0.381 \\
		ECO-HC & 0.584 & 0.524 & 0.572 & \textcolor[rgb]{ 0,  1,  0}{\textbf{0.599}} & 0.56  & 0.434 & 0.409 & 0.426 & \textcolor[rgb]{ 0,  1,  0}{\textbf{0.423}} & 0.421 \\
		Staple\_CA & 0.476 & 0.465 & 0.534 & 0.484 & 0.486 & 0.353 & 0.353 & 0.387 & 0.346 & 0.36 \\
		Staple & 0.463 & 0.498 & 0.567 & 0.491 & 0.512 & 0.343 & 0.379 & 0.407 & 0.349 & 0.377 \\
		KCF   & 0.376 & 0.31  & 0.38  & 0.363 & 0.353 & 0.251 & 0.227 & 0.262 & 0.242 & 0.24 \\
		SRDCF & 0.526 & 0.549 & 0.587 & 0.509 & 0.55  & 0.403 & 0.418 & 0.43  & 0.365 & 0.418 \\
		SAMF  & 0.414 & 0.36  & 0.427 & 0.418 & 0.391 & 0.293 & 0.267 & 0.303 & 0.288 & 0.281 \\
		\toprule[1.5pt]
	\end{tabular}%
	\label{tab:att_re}%
\end{table}%

\subsubsection{All-Day Performance}
To evaluate the abilities of the trackers in all-day tracking scenes, this part takes the average results on 3 benchmarks by sequences, \textit{i.e.} daytime benchmark UAV123@10fps \cite{Mueller2016ECCV}, DTB70 \cite{Li2017AAAI}, and nighttime benchmark UAVDark135, together 328 sequences. TABLE~\ref{tab:AVE} exhibits the average results of SOTA handcrafted CF-based trackers. Obviously, the proposed ADTrack possesses great advantages over all the other trackers in both DP and AUC. In specific, ADTrack (0.659) improves the precision of its baseline (0.581) by more than \textbf{13\%}, surpassing second place (0.651) by over \textbf{1\%}. In terms of AUC, ADTrack (0.480) is far ahead of the second place (0.468), up nearly \textbf{2.6\%}. Fig.~\ref{fig:vis} displays some representative tracking scenes in all-day condition, where ADTrack exhibits competitive in robustness against other arts. In addition to the satisfying tracking performance, ADTrack achieves an average speed of over 31 FPS on a single CPU, meeting real-time requirement of onboard UAV tracking.

Evidently, ADTrack achieves promising tracking performance in all-day condition, through day and night, thus greatly expanding the tracking based application onboard UAV around-the-clock.

\subsection{Attribute-Based Evaluation}\label{p3}
To clearly evaluate the abilities of the tracker under UAV specific challenge, this part displays their performance in the aforementioned 5 UAV tracking attributes, \textit{i.e.}, VC, FM, LR, OCC, and IV. TABLE~\ref{tab:att_re} gives the average results on all 328 sequences on 3 benchmarks. For the authoritative daytime benchmarks, this subsection follows our precious work \cite{Fu2020GRSM} to rewrite their official attributes. Clearly, ADTrack outperforms all other trackers in most attributes under 2 evaluation metrics. Specially, in FM, ADTrack surpasses second-best tracker ARCF-HC \cite{Huang2019ICCV} by over \textbf{3\%} in both DP and AUC. The results demonstrate the satisfying comprehensive tracking performance and favorable robustness of ADTrack in common challenges.

\subsection{Against Deep Trackers}\label{p4}
\begin{figure}[!t]
	\begin{center}
		\subfigure{
			\begin{minipage}{0.45\textwidth}
				\centering
				\includegraphics[width=1\columnwidth]{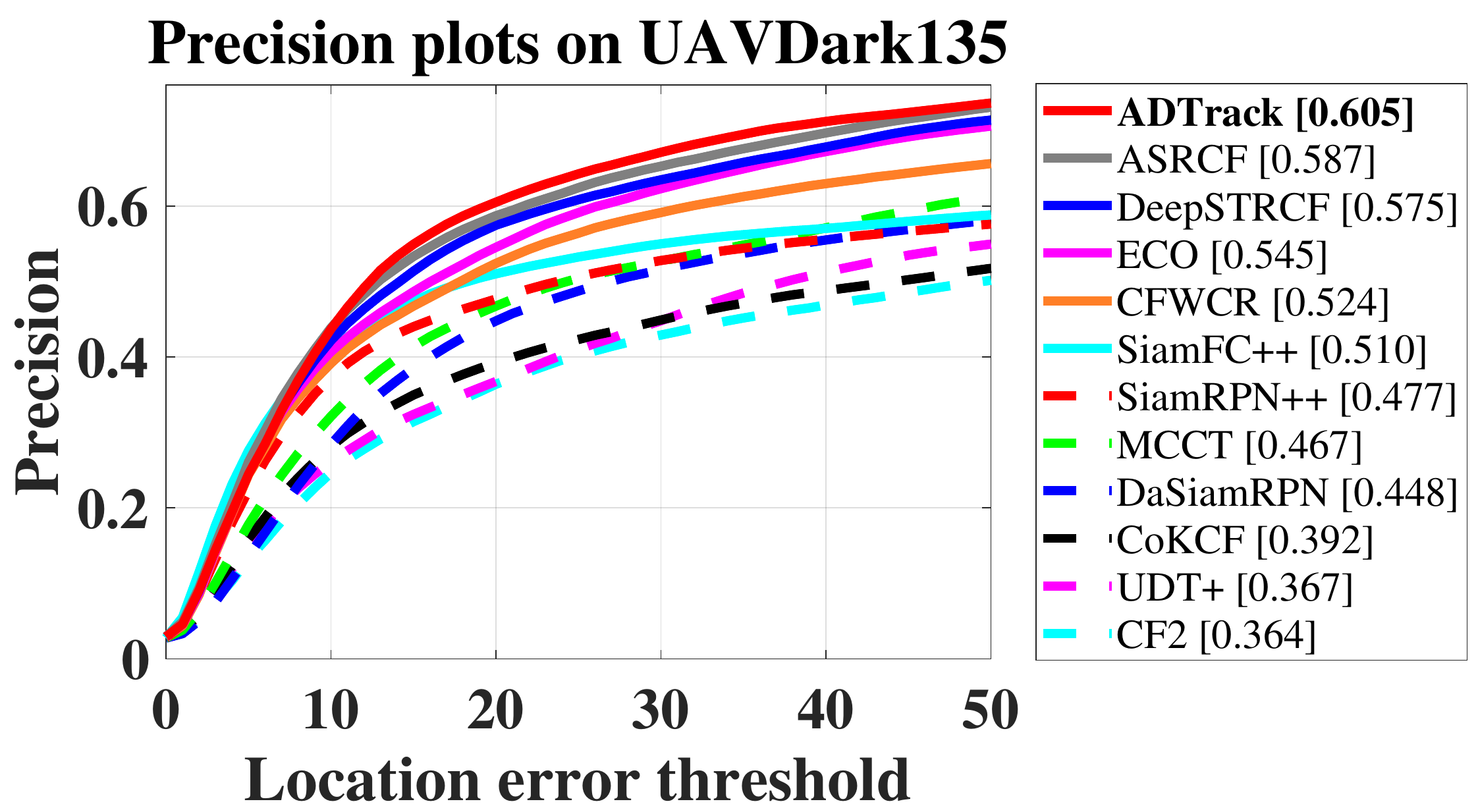}
			\end{minipage}
		}
	\\
		\subfigure{
			\begin{minipage}{0.45\textwidth}
				\centering
				\includegraphics[width=1\columnwidth]{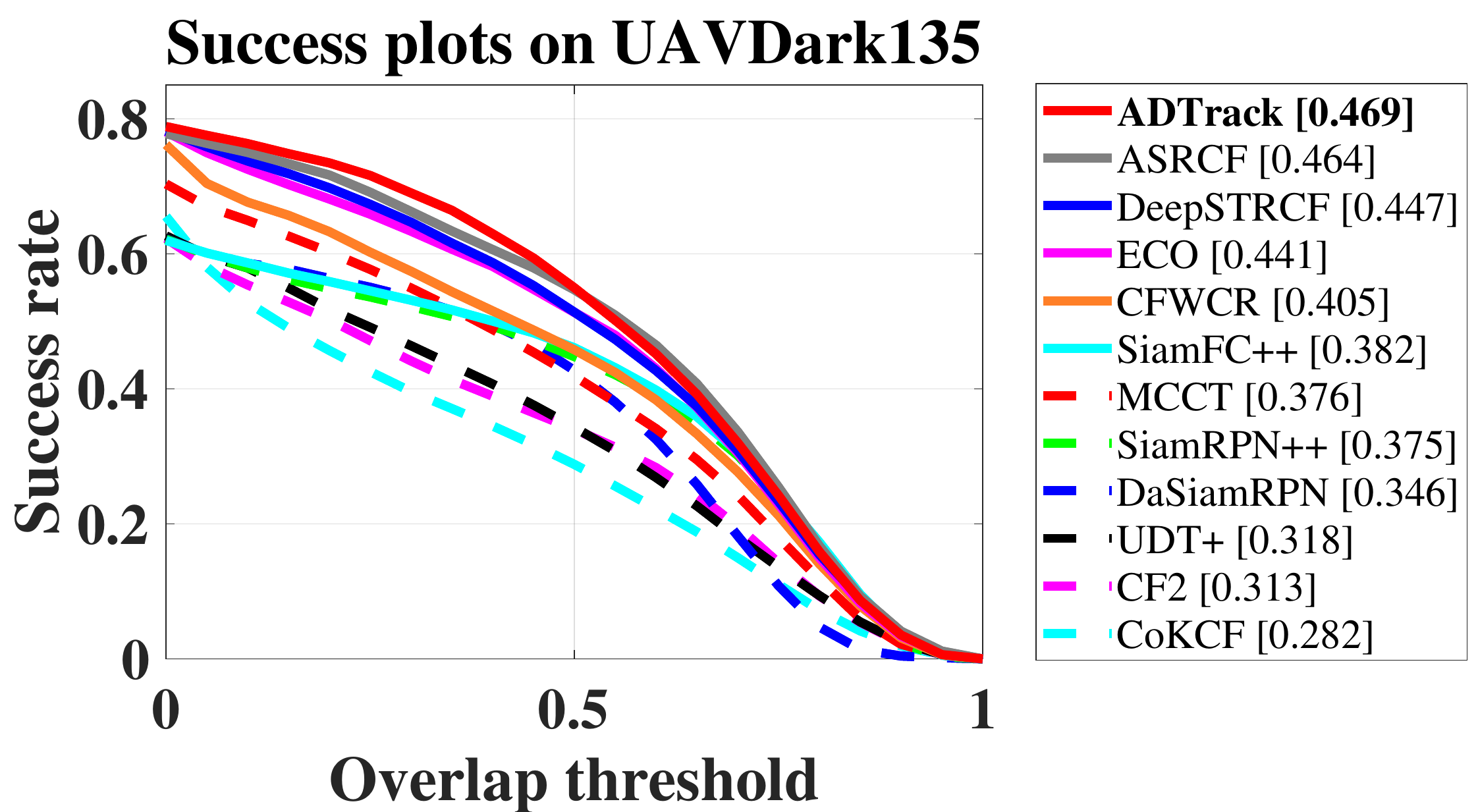}
			\end{minipage}
		}
	\end{center}
	\caption{Comparison of proposed ADTrack and the deep trackers on self-constructed UAVDark135. Evidently, most deep trackers fall short in hash dark scenes, while ADTrack maintains its robustness.}
	\label{fig:deep}
	\vspace{-0.1cm}
\end{figure}

This subsection focuses on comparison between proposed ADTrack and deep trackers which utilize off-line trained deep network for feature extraction or template matching. This work invites totally 11 SOTA deep trackers, \textit{i.e.}, SiamRPN++ \cite{Bo2019CVPR}, DaSiamRPN \cite{Zhu2018ECCV}, SiamFC++ \cite{Xu2020AAAI}, ASRCF \cite{Dai2019CVPR}, ECO \cite{Danelljan2017CVPR}, UDT+ \cite{wang2019CVPR}, HCFT \cite{ma2015ICCV}, CoKCF \cite{Zhang2017PR}, CFWCR \cite{He2017CVPR}, DeepSTRCF \cite{Li2018CVPR}, and MCCT \cite{wang2018CVPR}, to evaluate their performance in UAVDark135. From Fig.~\ref{fig:deep}, ADTrack outperforms all deep trackers in terms of DP and AUC under benchmark UAVDark135.

Using merely single CPU, ADTrack still achieves a real-time speed at over 30 FPS, while many deep trackers are far from real-time even on GPU, demonstrating the excellence of ADTrack for real-time UAV tracking against the deep trackers.

\Remark The results illustrate that the top-ranked deep trackers in recent years, especially the trackers without online update, \textit{e.g.}, SiamRPN++ \cite{Bo2019CVPR}, DaSiamRPN \cite{Zhu2018ECCV}, SiamFC++ \cite{Xu2020AAAI}, fail to maintain their robustness in real-world common dark scenes, since the off-the-shelf CNNs they utilize are trained by daytime images, ending up in their huge inferiority compared with online-learned ADTrack in the dark. Since there lacks sufficient dark images for training, off-line trained deep trackers fall short onboard UAV tracking at night.

\subsection{Ablation Study and Parameter Analysis}\label{p5}

\subsubsection{Component Validity Analysis}
\begin{figure}[!t]
	\centering
	\setlength{\abovecaptionskip}{-0.3cm}
	\includegraphics[width=1\columnwidth]{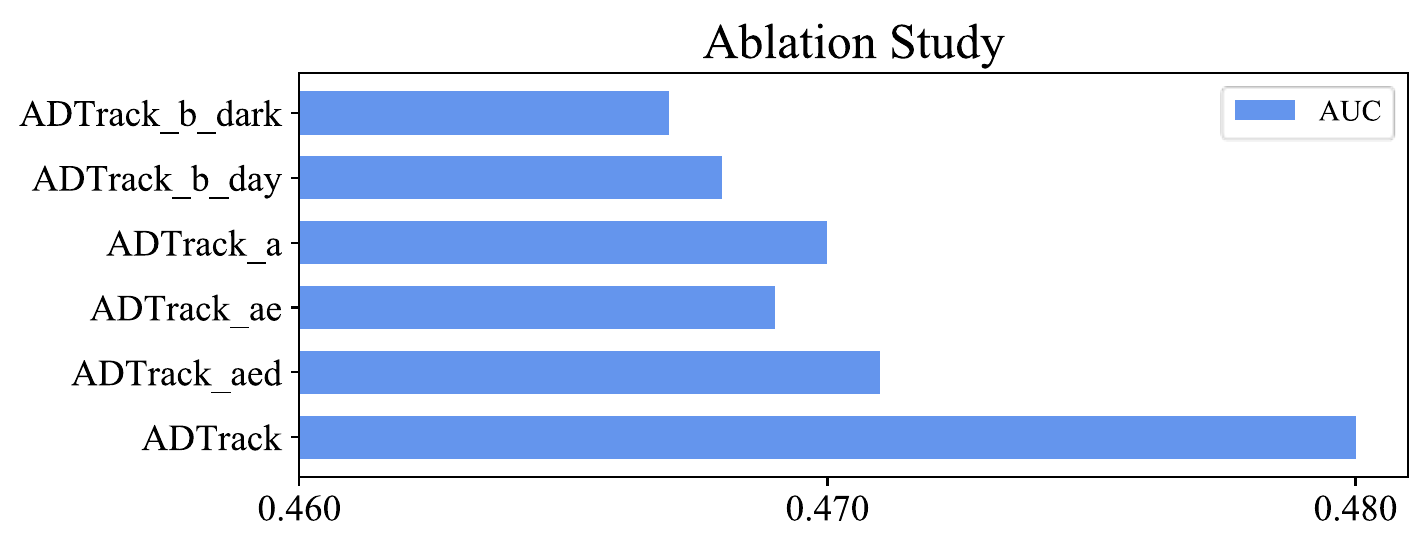}
	\caption{Ablation study of the modules in ADTrack. ADTrack, ADTrack\_aed, ADTrack\_ae, ADTrack\_a, ADTrack\_b\_day, and ADTrack\_b\_dark respectively denote ADTrack with different component activated. Note that ADTrack\_ae performs slightly inferior than ADTrack\_a since the enhancer needs to cooperate with the dual filter learning and dual response fusion modules to work effectively.
	}
	\label{fig:abla}
\end{figure}

To demonstrate the effectiveness of the proposed components, ablation study is conducted on all three benchmarks. The average AUC results by sequences on 3 benchmarks are displayed in Fig.~\ref{fig:abla}, where from bottom to top, proposed components, \textit{i.e.}, weight sum, dual filter constraint, and illumination adaptation, were disabled one by one. ADTrack\_aed denotes ADTrack without the weight sum in detection phase. In ADTrack\_ae, dual filter training constraint is disabled on the basis of ADTrack\_aed. ADTrack\_a denotes ADTrack\_ae without the image enhancer. And ADTrack\_b\_day, ADTrack\_b\_dark respectively denotes ADTrack with daytime parameters and nighttime parameters without any proposed module. The first 3 bars in Fig.~\ref{fig:abla} has demonstrated the validity of illumination adaptation module, and the last 3 bars illustrates how dual filter constraint and weight sum boosts the trackers' performance.

\subsubsection{Impacts of Key Parameters}
\begin{figure}[!t]
	\centering
	\setlength{\abovecaptionskip}{-0.3cm}
	\includegraphics[width=1\columnwidth]{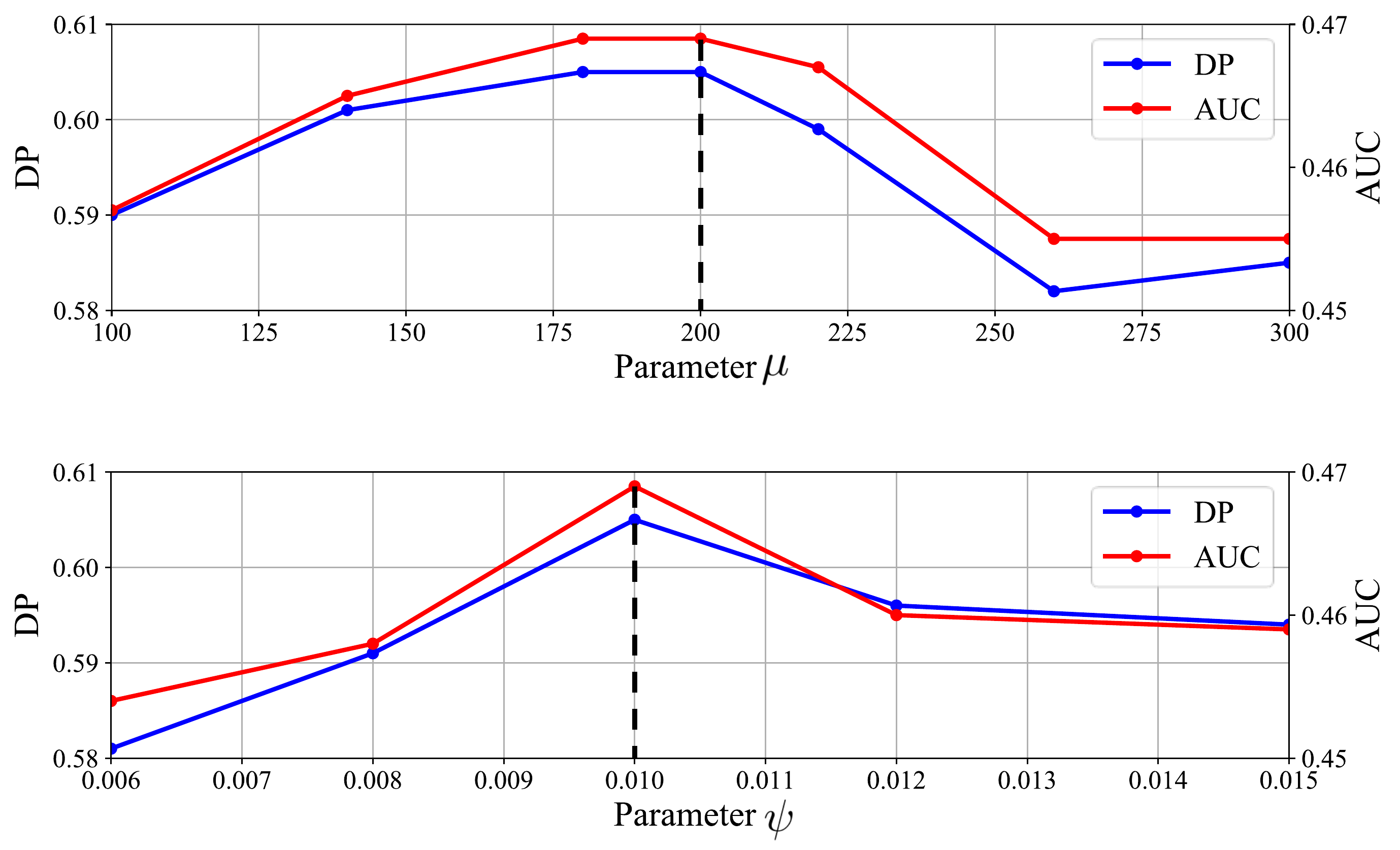}
	\caption{Parameter analysis of ADTrack on newly built benchmark UAVDark135. With other parameters remaining fixed, the tracking performance with different $\mu$ (Top) and with different $\psi$ (Bottom) are displayed. The chosen parameters along with their results are marked out by dotted lines.
	}
	\label{fig:para}
\end{figure}

The key parameters in ADTrack are the constraint parameter $\mu$ in the training regression equation and weight parameter $\psi$ in detection stage. We investigate the impact of two parameters on tracking results on nighttime benchmark UAVDark135, \textit{i.e.}, DP and AUC, which is shown in Fig.~\ref{fig:para}. Note that when $\mu$ varies, $\psi$ remains 0.01 unchanged, and when $\psi$ changes, $\mu$ is settled as 200.


\section{Conclusion}\label{Sec:6}
This work puts forward a novel real-time tracker with illumination adaptation and anti-dark function, \textit{i.e.}, ADTrack. ADTrack first implements illumination adaptation to decide day-night condition and switches its tracking mode. Then, pretreatment is carried out where proper training patch and target-aware mask are generated based on an image enhancer. With the mask, ADTrack proposes innovative dual filter regression model, in which the dual filters restrict each other in training and compensate each other in detection. In addition, the first large-scale dark tracking benchmark, UAVDark135, is also built in this work for visual tracking community. We strongly believe that the proposed tracker and dark tracking benchmark will make outstanding contribution to the research of UAV tracking in all-day conditions.

\section*{Acknowledgment}
This work is supported by the National Natural Science Foundation of China (No. 61806148) and Natural Science Foundation of Shanghai (No. 20ZR1460100).

\ifCLASSOPTIONcaptionsoff
  \newpage
\fi

\bibliographystyle{IEEEtran}

\bibliography{TIP}

\end{document}